\documentclass[10pt]{article} 
\usepackage[preprint]{rlc}

\usepackage{amssymb}            
\usepackage{mathtools}          
\usepackage{mathrsfs}           
\mathtoolsset{showonlyrefs}     
\usepackage{graphicx}           
\usepackage{subcaption}         
\usepackage[space]{grffile}     
\usepackage{url}                
\usepackage{svg}

\usepackage{xspace}

\makeatletter
\DeclareRobustCommand\onedot{\futurelet\@let@token\@onedot}
\def\@onedot{\ifx\@let@token.\else.\null\fi\xspace}

\def\ie{\emph{i.e}\onedot}

\makeatother

\let\classAND\AND
\let\AND\relax
\usepackage{algorithmic}

\let\AND\classAND
\AtBeginEnvironment{algorithmic}{\let\AND\algoAND}

\usepackage[textsize=tiny]{todonotes}

\title{Enhancing Policy Gradient with the Polyak Step-Size Adaption}

\author{Yunxiang Li  \\
    yunxiang.li@mbzuai.ac.ae \\
    \And
    Rui Yuan \thanks{The work was done when the author was affiliated with Télécom Paris.} \\
    rui.yuan@stellantis.com
    \And
    Chen Fan \\
    fanchen3@outlook.com \\
    \And
    Mark Schmidt \\
    schmidtm@cs.ubc.ca \\
    \And
    Samuel Horv\'ath \\
    samuel.horvath@mbzuai.ac.ae \\
    \And
    Robert M. Gower \\
    gowerrobert@gmail.com \\
    \And
    Martin Taká\v{c} \\
    martin.takac@mbzuai.ac.ae
    }

\begin{document}

\maketitle


\begin{abstract}
Policy gradient is a widely utilized and foundational algorithm in the field of reinforcement learning (RL). Renowned for its convergence guarantees and stability compared to other RL algorithms, its practical application is often hindered by sensitivity to hyper-parameters, particularly the step-size. In this paper, we introduce the integration of the Polyak step-size in RL, which automatically adjusts the step-size without prior knowledge. To adapt this method to RL settings, we address several issues, including unknown $f^*$ in the Polyak step-size. Additionally, we showcase the performance of the Polyak step-size in RL through experiments, demonstrating faster convergence and the attainment of more stable policies.
\end{abstract}

\section{Introduction} \label{sec:intro}

The policy gradient serves as an essential algorithm in various cutting-edge reinforcement learning (RL) techniques, such as natural policy gradient (NPG), TD3, and SAC \citep{DBLP:conf/nips/Kakade01, DBLP:conf/icml/FujimotoHM18, DBLP:conf/icml/HaarnojaZAL18}. Such method computes the gradient for the policy performance measurement and applies gradient ascent to enhance performance. Recognized for its outstanding performance in control tasks and convergence guarantees, the policy gradient has garnered considerable attention. However, similar to many RL algorithms, it exhibits a high sensitivity to hyper-parameters \citep{DBLP:conf/icml/EimerLR23}. The challenge is exacerbated by the time-consuming nature of hyper-parameter tuning, particularly in large-scale tasks.

\begin{figure}[thb] \centering
    \includegraphics[width=0.4\textwidth]{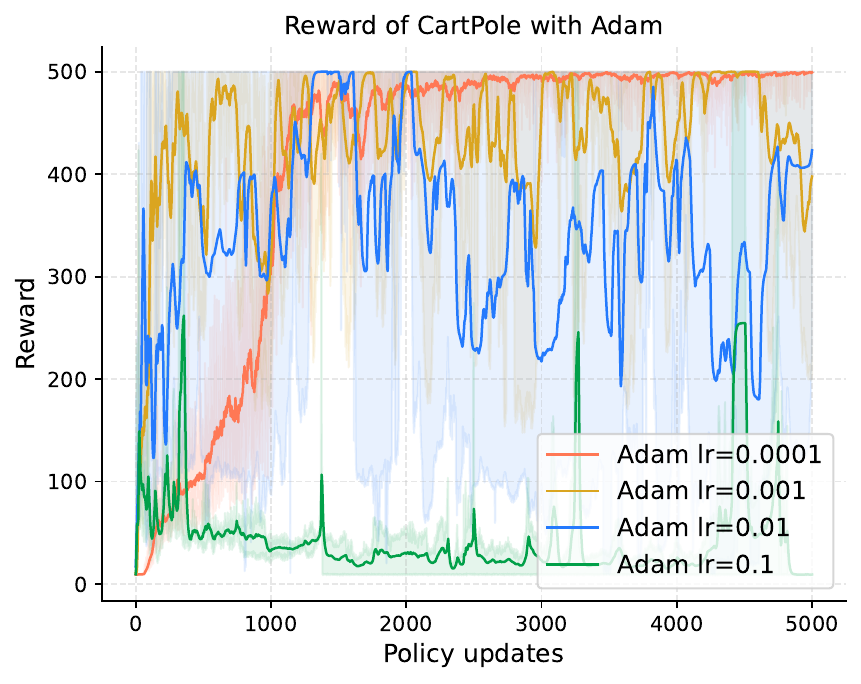}
    \includegraphics[width=0.4\textwidth]{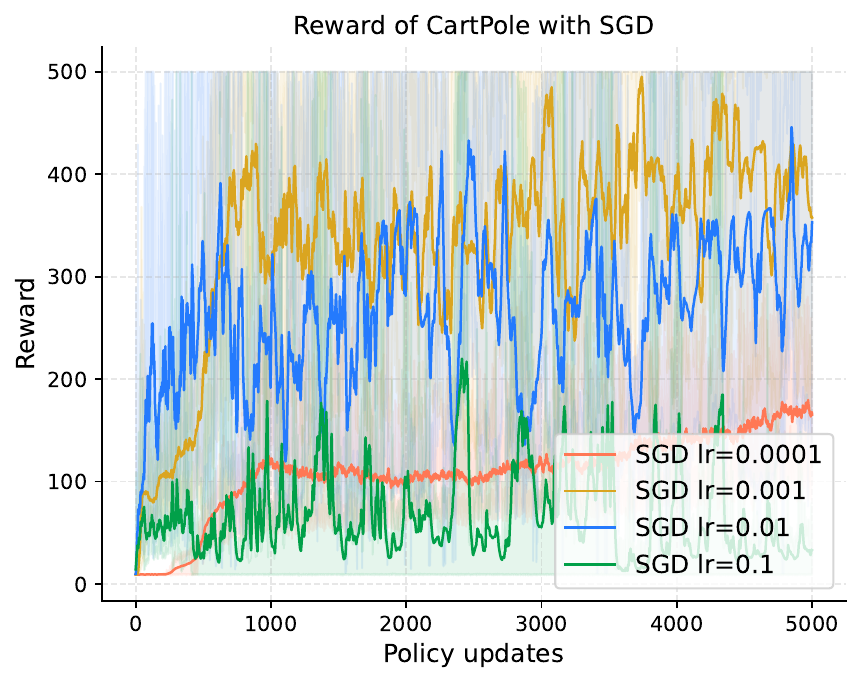}
    \caption{The performance of Adam and SGD with various step-sizes evaluated on the CartPole environment \cite{brockman2016openai}. The evaluation is averaged across three unique random seeds, distinct from the training seeds. The reported evaluation rewards are presented as a moving average due to oscillations.} 
    \label{fig:adam_sgd_lr}
\end{figure}

In this paper, our focus lies on the step-size (or learning rate) in policy gradient, a parameter that significantly influences algorithm performance \citep{DBLP:conf/icml/EimerLR23}. Moreover, given the variation in reward scales and landscapes across different tasks \citep{DBLP:journals/corr/abs-2009-02391}, fine-tuning learning parameters from one task for application in another is impractical. Instead, the optimal step-size needs to be determined independently for each task.
Traditionally, the prevailing approach for selecting a step-size involves evaluating a range of plausible values and choosing a fixed value that leads to the fastest convergence to the optimal solution. However, this method is computationally expensive.
An alternative practical choice is the use of diminishing step sizes, which may offer improved convergence speed and sample efficiency empirically. Nevertheless, the effectiveness of this approach relies heavily on intuition and experience. Similar to fixed step-sizes, determining good practices for one environment is challenging when transferred to another environment.
We illustrate the impact of step-size on performance and sample efficiency with Adam \citep{DBLP:journals/corr/KingmaB14} and stochastic gradient descent (SGD) in Figure \ref{fig:adam_sgd_lr}. Notably, with Adam, a larger step-size reaches the local optimum faster, while a smaller step-size leads to a stable policy over time. With SGD, achieving both stability and an optimal policy is shown to be more challenging.

To address this challenge, we propose an adaptive step-size method for policy gradient. Drawing inspiration from the Polyak step-size concept \citep{Poliak1987IntroductionTO}, we tailor it for application in policy gradient, especially stochastic update. Unlike traditional approaches with sensitive step-sizes, our method computes the step-size for each policy update step using robust hyper-parameters. We validate the effectiveness of our approach through numerical experiments and showcase its performance compared to the widely used Adam algorithm \citep{DBLP:journals/corr/KingmaB14}. 
The key contributions of our work are outlined below:
\begin{itemize}
    \item {\bf Adoption of the Polyak Step-Size Idea:} We integrate the Polyak step-size concept into the policy gradient framework, eliminating the need for sensitive step-size fine-tuning by employing robust hyper-parameters.
    \item {\bf Investigation and Resolution of Issues:} We systematically investigate and address the challenges associated with applying Polyak step-size to policy gradient, ensuring its practicality and effectiveness.
    \item {\bf Demonstrated Performance:} Through experiments, we provide empirical evidence that our proposed method outperforms alternative approaches, showcasing its efficacy in RL tasks. Our method excels in terms of faster convergence, enhanced sample efficiency, and the maintenance of a stable policy.
\end{itemize}

\section{Related work}
In response to the success of SGD and its central dependency on step-size for convergence, numerous works have focused on optimizing this parameter. The Stochastic Polyak Step-size \citep[][SPS]{DBLP:conf/aistats/LoizouVLL21} and Stochastic Line-Search  \citep[][SLS]{DBLP:conf/nips/VaswaniMLSGL19} introduced Polyak step-size and line-search for interpolation tasks, inspiring various subsequent variants. Noteworthy contributions include \citet{DBLP:conf/icml/VaswaniDB22}'s validation of an exponentially decreasing step-size with SLS, adapting well to noise and problem-dependent constants. \citet{DBLP:conf/nips/OrvietoLL22} extended SPS to non-over-parameterized. \citet{DBLP:journals/corr/abs-2308-06058} further solved the problems in non-interpolation settings and guaranteed the convergence rate. \citet{DBLP:journals/corr/abs-2305-07583} proposed momentum-based adaptive learning rates.

While the step-size plays a crucial role in determining the performance and reproducibility of RL algorithms, only a limited number of works have addressed the challenge of sensitive step-size tuning in RL.
\citet{DBLP:journals/jmlr/MatsubaraMM10} extended the metric in natural policy gradient (NPG) and introduced the Average reward metric Policy Gradient method (APG). NPG measures the impact on the action probability distribution concerning the policy parameters, while APG introduces an average reward metric with a Riemannian metric, directly evaluating the effect on the average reward of policy improvement. The authors present two algorithms incorporating the average reward metric as a constraint.
\citet{DBLP:conf/nips/PirottaRB13} derived a lower bound for the performance difference in the context of full-batch policy gradient, demonstrating a fourth-order polynomial relationship with the step-size. They extend their analysis to Gaussian policies, providing a quadratic form of the step-size in the lower bound. By obtaining a closed form for the maximum, they ensure monotonic improvement in each iteration. Additionally, they offer a simplified version that guarantees improvement with high probability using sample trajectories.
\citet{DBLP:conf/aaai/DabneyB12} investigated step-size in the temporal difference learning problem with linear approximation. They derived an adaptive tight upper bound for the optimal step-size and presented a heuristic step-size computation method with efficient computation and storage.
In contrast to manual tuning, \citet{DBLP:conf/icml/EimerLR23} conducted a comprehensive study across state-of-the-art hyper-parameter optimization techniques, and provided easy-to-use implementations for practical use, while it requires the knowledge of problem-dependent constants.

Despite the great ideas in previous work, our contribution not only provides adaptive learning rates across various tasks without requiring the knowledge of any problem-dependent constants, but also showcases better convergence speed. In experiments, we observe that our step-size is larger than the typically used constant step-size with Adam. Unlike previous pure theory works, our conducted experiments serve to validate and ensure the reproducibility of our findings.

\section{Preliminaries}
Let $\mathcal{M} = (\mathcal{S}, \mathcal{A}, \rho, P, r, \gamma)$ represent a discounted Markov Decision Process (MDP) \citep{puterman2014markov} with state space $\mathcal{S}$, action space $\mathcal{A}$, initial state distribution $\rho$, transition probability function $P(s^\prime \mid s, a): \mathcal{S}\times \mathcal{A}\times\mathcal{S} \rightarrow [0, 1]$, reward function $r: \mathcal{S}\times \mathcal{A} \rightarrow \mathbb{R}$, and discounted factor $\gamma \in [0, 1]$. 
A policy on an MDP is defined as a mapping function $\pi \in  \Delta(\mathcal{A})^{\mathcal{S}}$. A trajectory, denoted as $\tau = (s_0, a_0, r_0, s_1, a_1, r_1, \dots)$, is generated by following the transition function, the reward function, and the policy, with $r_t = r(s_t, a_t)$.

Given a policy $\pi$, let $V^\pi: \mathcal{S}\rightarrow \mathbb{R}$ denote the value function of state $s$ defined as
\begin{align*}
    V^\pi (s) = \int_\tau P(\tau \mid s_0 = s) R(\tau) d\tau
    = \mathbb{E}_{\tau} \left[\sum_{t=0}^\infty \gamma^t r_t \mid s_0 = s \right],
\end{align*}
where $P(\tau\mid s_0=s) = \rho(s)\prod_{t=0}^\infty \pi(a_t \mid s_t) P(s_{t+1} \mid s_t, a_t)$ represents the probability of the trajectory following the transition function, the reward function, and the policy, $R(\tau) = \sum_{t=0}^\infty \gamma^t r_t$ is the discounted sum of rewards in the trajectory $\tau$.

Given the starting state distribution $\rho$, our objective is to find a policy that maximizes the objective function
\begin{equation*}
    V^\pi (\rho) = \mathbb{E}_{s \sim \rho(s)} \left[ V^\pi(s) \right].
\end{equation*}
In policy optimization, we also employ the useful definition of the $Q$ value for a state-action pair $(s, a) \in \mathcal{S} \times \mathcal{A}$, defined as
\begin{equation*}
    Q^\pi (s, a) = \mathbb{E}_{\tau} \left[\sum_{t=0}^\infty \gamma^t r(s_t, a_t) \mid s_0 = s, a_0 = a \right].
\end{equation*}

In practice, we use model parameter $\theta$ to define the policy $\pi^\theta$. In this paper, we analyze policies with softmax parametrizations. For simplicity, we omit $\pi$ in the notations and use $\theta$, namely, $V^\theta(s), V^\theta(\rho), Q^\theta(s, a)$. The gradient with respect to the model parameter $\theta$ is
\begin{equation} \label{eq:grad_V}
    \nabla_\theta V^\theta (\rho) = \int_s \rho(s) \int_\tau \nabla_\theta P(\tau \mid s_0 = s) R(\tau) d\tau ds.
\end{equation}
Then we update the parameters $\theta$ with the gradient
\begin{equation*}
    \theta_{t+1} = \theta_t + \eta_t \nabla_{\theta_t} V^{\theta_t} (\rho)
\end{equation*}
to maximize the objective function. To avoid symbol conflict, we use $\eta_t$ to represent the step size at time $t$. This general schema is called policy gradient \citep{williams1992simple,Sutton1998ReinforcementLA}.

\subsection{GPOMDP}

To handle the unknown transition function and intractable trajectories in the objective function and the gradient, a common approach is to sample trajectories and estimate. Specifically, for an estimate value function $\hat{V}^\theta$, we simulate $m$ truncated trajectories $\tau_i = \left(s_0^i, a_0^i, r_0^i, s_1^i, \cdots, s_{H-1}^i, a_{H-1}^i, r_{H-1}^i\right)$ with $r^i_t = r(s_t^i, a_t^i)$, and with horizon $H$ using the defined policy and compute the discounted sum of rewards. \citet{DBLP:journals/jair/BaxterB01} extended score function or likelihood ratio method \eqref{eq:grad_V} \citep[][REINFORCE]{williams1992simple} and introduced GPOMDP, generating the estimate of the gradient of the parameterized stochastic policies:
\begin{equation*}
\widehat{\nabla}_\theta V^\theta(\rho) \; = \; \frac{1}{m} \sum_{i=1}^m \sum_{t=0}^{H-1} 
\left( \sum_{t'=0}^t \nabla \log \pi^\theta(a_{t'}^i \mid s_{t'}^i) \right)\gamma^t r^i_t.    
\end{equation*}

\subsection{The Polyak step-size}
We first introduce the Polyak step-size \citep{Poliak1987IntroductionTO} within the optimization context, specifically when optimizing a finite-sum function $f(x) = \frac{1}{n}\sum_{i=1}^n f_i(x)$. The objective is to minimize $f(x)$ with the model parameter $x$. We denote $x^*$ as the optimal point and $f^*$ as the minimum value of $f$. This optimization problem can be addressed using gradient descent (GD) or SGD:
\begin{align*}
    x^{k+1} &= x^k - \gamma_k \nabla f(x^k), && \text{(GD)}  \\
    x^{k+1} &= x^k - \gamma_k \nabla f_i(x^k), && \text{(SGD)}
\end{align*}
where $\gamma_k > 0$ is the step-size at iteration $k$.

The classic Polyak step-size is defined as
\begin{equation*}
    \gamma_k = \frac{f(x^k) - f^*}{\|\nabla f^k\|^2},
\end{equation*}
which minimizes the following upper bound $Q(\gamma) = \|x^k - x^*\|^2 - 2 \gamma [f(x^k) - f^*] + \gamma^2 \|\nabla f(x^k)\|^2$ for convex functions. In a recent work, \citet{DBLP:conf/aistats/LoizouVLL21} extended the Polyak step-size to the stochastic setting, namely SPS$_{\max}$, 

\begin{equation*}
    \gamma_k = \min\left\{\frac{f_i(x^k) - f_i^*}{c\|\nabla f_i(x^k)\|^2}, \gamma_b\right\},
\end{equation*}
where $f_i^* = \inf_x f_i(x)$, $c$ is a positive constant, and $\gamma_b>0$ is to restrict SPS from being very large.

In RL settings, the goal is to maximize the objective function (\ie, the value function). To this end, we extend  $\operatorname{SPS}_\text{max}$ to
\begin{equation*}
    \gamma_k = \min\left\{\frac{V^* - \hat{V}^{\theta_k}}{c\|\nabla_{\theta_k} \hat{V}^{\theta_k}\|^2}, \gamma_b\right\},
\end{equation*}
where $V^*$ is the optimal objective function value, $\hat{V}^{\theta_k}$ is a stochastic evaluation of the current policy parameter $\theta_k$.

\section{Policy gradient with the Polyak step-size}

In the direct application of the Polyak step-size from optimization to RL, several issues arise, which we will detail in the following sections. We will then adapt the Polyak step-size to address these issues.

\subsection{Stochastic update issue} \label{sec:stoc_issue}

\begin{figure}[thb] \centering
    \includegraphics[width=0.3\textwidth]{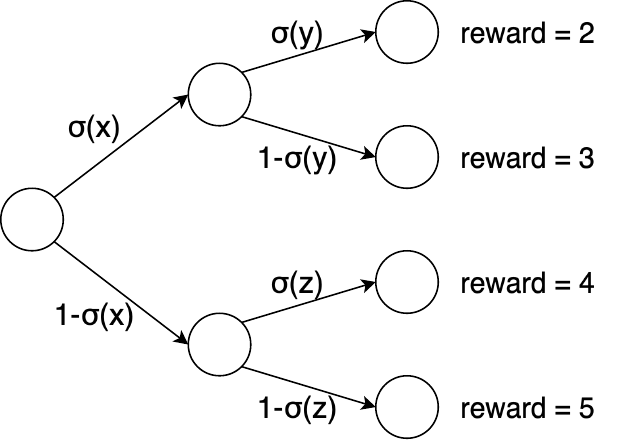}
    \caption{A simplified two-step deterministic environment with a three-parameter policy. $x$, $y$, and $z$ represent the parameters of the policy. The sigmoid function is denoted by $\sigma$ with $\sigma(u) = \frac{1}{1+e^{-u}}$. Selecting a non-optimal trajectory at the first iteration leads to an increase in the probability of such a trajectory. Consequently, the update with the Polyak step-size is likely to approach infinity.} \label{fig:2-step-env}
\end{figure}

First, in most real-world scenarios, trajectories and transition functions are not tractable, necessitating the adoption of stochastic sampling to estimate the objective function $V^\theta$ and the gradient $\nabla_\theta V^\theta$. To illustrate a problem arising from stochastic updates, we employ a simplified CartPole environment from OpenAI Gym \citep{brockman2016openai}, where the size of the action space is $2$.
We concentrate on the initial two steps and four trajectories (seven distinct states) within the CartPole environment. Employing a softmax parametrization, our policy is defined by three parameters ($x, y$ and $z$ described in Figure \ref{fig:2-step-env}). Distinct rewards are assigned to each trajectory, as illustrated in Figure \ref{fig:2-step-env}. The optimal policy is deterministic, consistently selecting the trajectory with the highest reward.

Suppose we initialize the parameters as zeros, indicating uniform trajectory sampling. We use a single sample trajectory to estimate both $V^\theta$ and the gradient $\nabla_\theta V^\theta$. If a non-optimal trajectory $\tau$ is sampled in the first iteration, the probability of the trajectory $P(\tau)$ is increased, because of \eqref{eq:grad_V} and the $R(\tau) > 0$. The same trajectory will probably be repeatedly selected in subsequent iterations, further increasing its probability. In a softmax policy, as the probability tends toward determinism, the gradient approaches zero. In non-optimal trajectories with nearly deterministic probabilities, the Polyak step-size formula encounters an issue: $V^* - V^\theta$ is not zero while $\frac{\nabla_\theta V^\theta}{\|\nabla_\theta V^\theta\|^2}$ approaches infinity, resulting in an explosive update. 
This issue extends beyond positive trajectory rewards; whenever a non-optimal trajectory has a high probability, the potential for explosive updates arises.

To mitigate this issue, we introduce an entropy penalty into the policy update loss, a practical remedy to keep the probabilities from getting too small \citep{williams1991function,peters10relative,volodymyr2016asynchronous,alfano2023novel}. This modification prevents $\frac{\nabla_\theta V^\theta}{\|\nabla_\theta V^\theta\|^2}$ from approaching infinity, promoting exploration and ensuring the likelihood of sampling optimal trajectories. Consequently, the objective function for $\theta$ becomes:
\begin{equation*}
    \mathcal{L}(\theta) = V^\theta(\rho) + \alpha \mathbb{E}_{s \sim d^{\pi^\theta} (s)} \mathcal{H}(\pi^\theta(\cdot \mid s)),
\end{equation*}
where $\alpha$ is the hyper-parameter which determines the importance of the entropy, $\mathcal{H}(\pi^\theta(\cdot \mid s)) = \sum_{a \in \mathcal{A}} -\pi^\theta(a \mid s)\log\pi^\theta(a \mid s)$ is the entropy function, $d^{\pi^\theta}(s) = (1-\gamma) \mathbb{E}_{s_0 \sim \rho}\sum\nolimits_{t=0}^\infty \gamma^t \Pr[s_t = s \mid s_0, \pi^\theta]$ is the normalized discounted state visitation distribution which adjusts the weights of entropy. Intuitively, the state visitation distribution measures the probability of being at state $s$ across the entire trajectory.
The entropy term is also intractable, we estimate it with sample trajectories.

\subsection{Estimating $V^*$ with twin-model method}

Computing the Polyak step-size requires knowledge of the optimal objective function values. Since the $V^\theta(\rho)$ is non-concave in $\theta$ for the softmax parameterizations \citep{Agarwal2019ReinforcementLT,agarwal2021theory}, using the true $V^*$ value could easily cause large step-size and overshooting. Additionally, in real situations, we may not know the true $V^*$ value, as we lack information about optimal policies or because of the nature of the task. To address the $V^*$ problem, we propose a twin-model method. 

We first introduce our method within the context of stochastic optimization, minimizing $f(x) = \frac{1}{n}\sum_{i=1}^n f_i(x)$. We initialize two distinct model parameters, denoted as $x_1$ and $x_2$. At each iteration, we compute $f_i(x_1)$ and $f_j(x_2)$,
where $i$ and $j$ are uniformly sampled with replacement \footnote{We can also sample $i$ first and then sample $j$ until $j \neq i$ (sampling without replacement), in this case $\mathbb{E}_j[f_j(x)] = f(x)$ also holds.},
select the lower value as $f^*$ and update the other model. 

For instance, at $k$-th iteration, assuming $f_i(x_1^k) < f_j(x_2^k)$, the expression for the Polyak step-size is then given by:
\begin{equation*}
    \gamma_k = \frac{f_j(x_2^k) - f_i(x_1^k)}{\|\nabla f_j(x_2^k)\|^2},
\end{equation*}
and we freeze the parameter $x_1^k$ and update the parameter $x_2^k$ with:
\begin{equation*}
    x_2^{k+1} = x_2^k - \gamma_k \nabla f_j(x_2^k).
\end{equation*}
When $f_i(x_1^k) = f_j(x_2^k)$, we don't update the models and start a new iteration.
In practical applications, our findings suggest that the stochastic update prevents the two models from converging closely. To ensure similar performance for a conservative update, we initialize the models closely.

As in the RL setting, adopting two policy models for better value estimation is a common practice \citep{Hasselt2010DoubleQ, DBLP:journals/corr/LillicrapHPHETS15, DBLP:conf/icml/HaarnojaZAL18, DBLP:conf/icml/FujimotoHM18}.
Since we usually use a deep neural network as model parameter $\theta$, one option is to use two initializations of the same structure, denoted as $\theta_1$ and $\theta_2$. We evaluate them with stochastic trajectory samples in each iteration. Due to different parameters and likely different stochastic samples, we obtain distinct estimations $\hat{V}^{\theta_1}$ and $\hat{V}^{\theta_2}$. We use the higher $\hat{V}$ to update the model with the lower $\hat{V}$. For instance, in iteration $k$, if $\hat{V}^{\theta_{1}^k} > \hat{V}^{\theta_{2}^k}$, we freeze $\theta_{1}^k$ and update $\theta_{2}^k$ with a step-size:
\begin{equation*}
\hat{\gamma}_k = \frac{\hat{V}^{\theta_{1}^k} - \hat{V}^{\theta_{2}^k}}{\|\widehat{\nabla}_{\theta_{2}^k} V^{\theta_{2}^k}\|^2}.
\end{equation*}

This constitutes a pessimistic estimation of $V^*$, closely aligned with the current $\hat{V}^{\theta_{2_k}}$. In comparison with the step-size involving the true $V^*$, $\hat{\gamma}_k$ is smaller and more conservative, reducing the likelihood of too large step-size. 
Due to the issue raised in Section \ref{sec:stoc_issue}, it is still likely to sample non-optimal trajectories which has high probability with the current policy. Thus in practice, we use the bounded stochastic variant $\operatorname{SPS}_\text{max}$:
\begin{equation*}
    \hat{\gamma}_k = \min\left\{\frac{\hat{V}^{\theta_{1}^k} - \hat{V}^{\theta_{2}^k}}{c\|\widehat{\nabla}_{\theta_{2}^k} V^{\theta_{2}^k}\|^2}, \gamma_b \right\}.
\end{equation*}

\subsection{Algorithm}
Combining the twin-model method for estimating $V^*$ and the entropy penalty, where we use $\hat{\mathcal{L}}(\theta)$ instead of $\hat{V}^\theta$, we propose our algorithm as outlined in Algorithm \ref{alg:algo}.
\begin{algorithm}
   \caption{Double policy gradient with the stochastic Polyak step-size}
   \label{alg:algo}
\begin{algorithmic}
   \STATE {\bfseries Input:} $\gamma_b$, $c$, $\alpha$, different but close model initializations $\theta_1$ and $\theta_2$
   \REPEAT
   \STATE Sample trajectories $\{\mathcal{T}_1\}$ and $\{\mathcal{T}_2\}$ with $\theta_1$ and $\theta_2$ accordingly
   \STATE Compute $\hat{\mathcal{L}}(\theta_1)$ and $\hat{\mathcal{L}}(\theta_2)$ with $\{\mathcal{T}_1\}$ and $\{\mathcal{T}_2\}$
   \IF{$\hat{\mathcal{L}}(\theta_1) > \hat{\mathcal{L}}(\theta_2)$}
   \STATE Compute $\widehat{\nabla}_{\theta_2} \mathcal{L}(\theta_2)$ with $\{\mathcal{T}_2\}$
   \STATE $\gamma = \min\{\frac{\hat{\mathcal{L}}(\theta_1) - \hat{\mathcal{L}}(\theta_2)}{c\|\widehat{\nabla}_{\theta_2} \mathcal{L}(\theta_2)\|^2}, \gamma_b\}$
   \STATE $\theta_2 \leftarrow \theta_2 + \gamma \widehat{\nabla}_{\theta_2} \mathcal{L}(\theta_2)$
   \ELSE
   \STATE Compute $\widehat{\nabla}_{\theta_1}\mathcal{L}(\theta_1)$ with $\{\mathcal{T}_1\}$
   \STATE $\gamma = \min\{\frac{\hat{\mathcal{L}}(\theta_2) - \hat{\mathcal{L}}(\theta_1)}{c\|\widehat{\nabla}_{\theta_1} \mathcal{L}(\theta_1)\|^2}, \gamma_b\}$
   \STATE $\theta_1 \leftarrow \theta_1 + \gamma \widehat{\nabla}_{\theta_1} \mathcal{L}(\theta_1)$
   \ENDIF
   \UNTIL{$\gamma$ is small enough}
\end{algorithmic}
\end{algorithm}

\section{Experiments}

\subsection{twin-model in optimization problems}
\begin{figure}[htbp!]
    \centering
      \begin{subfigure}{0.45\textwidth}
        \includegraphics[width=\textwidth]{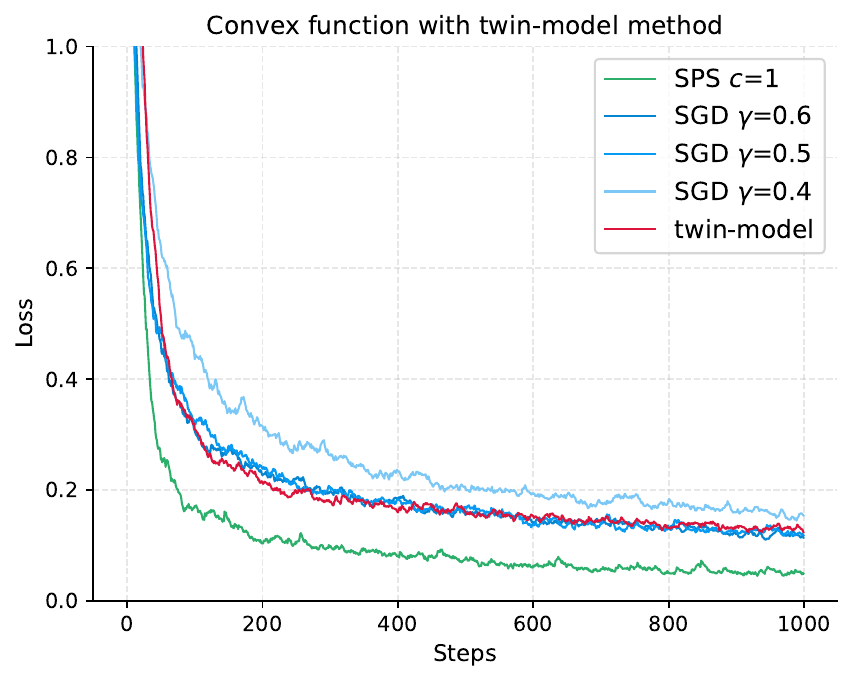}
        \caption{}
        \label{fig:convex}
      \end{subfigure}
      \hfill
      \begin{subfigure}{0.45\textwidth}
        \includegraphics[width=\textwidth]{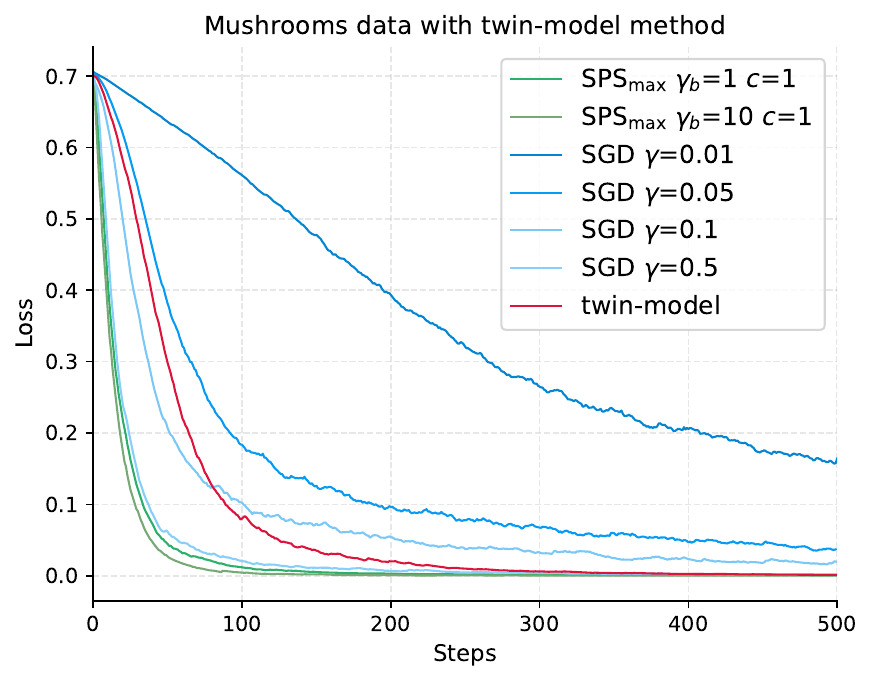}
        \caption{}
        \label{fig:non-convex}
      \end{subfigure}
\caption{The performance of twin-model method on convex and non-convex problem comparing with SGD with various step-sizes and SPS (SPS$_\text{max}$).}
\end{figure}

We showcase the effectiveness of the twin-model method under interpolation conditions for simple convex function minimization, specifically logistic regression (\ie, the data is linearly separable). We apply simple $\operatorname{SPS}$ with $c=1$.
In this case, we remove the requirements of $f^*$, $c$ and $\gamma_b$, which means we require no hyper-parameters and no prior knowledge.
The performance comparison involves the Polyak step-size with the twin-model method, SGD with varying step-sizes and SPS, as illustrated in Figure \ref{fig:convex}. Our evaluation incorporates three distinct random seeds, and we present the moving average of $f(x)$. Notably, the twin-model method exhibits similar performance to fine-tuned SGD in addressing convex problems under interpolation conditions.
Under identical experimental settings, \ie, no prior knowledge and no hyper-parameters available for twin-model method, we conduct a comparison between the twin-model method, SGD with different step-sizes and SPS$_\text{max}$ on non-convex problem (\ie, logistic regression with non-convex deep neural network). The results are presented in Figure \ref{fig:non-convex}. Notably, the twin-model method demonstrates comparable performance to fine-tuned SGD.
In the RL setting, the main challenge lies in the uncertainty surrounding $V^*$ ($f^*$ in optimization settings), which may not be known.

\subsection{Experiments settings}
\paragraph{Training.}
In this section, we elaborate on the details of our experiments. We establish the baseline using the Adam optimizer as provided in PyTorch \citep{paszke2017automatic}. For gradient estimation, we employ the GPOMDP method, utilizing $500$ trajectories for each policy update. Our investigation includes variations of Adam with different fixed step-sizes, both with and without the incorporation of an entropy penalty.

For the Polyak step-size, we collect $500$ trajectories for each of $\theta_1$ and $\theta_2$. We use the average of the sampled trajectory rewards as $V$ and apply GPOMDP to approximate the policy gradient. Our investigation encompasses various values for the $c$ in $\operatorname{SPS}_\text{max}$ and the upper bound of the step-size $\gamma_b$. Both the ordinary policy gradient and our proposed method employ fully connected feed-forward neural networks with $128$ neurons in the hidden layer.


\paragraph{Evaluation.}
Since there exists a stationary deterministic policy which is an optimal policy\citep{Agarwal2019ReinforcementLT}, 
we select the action with the highest probability from the policy in evaluation. After each policy update, we evaluate the policies with three different random seeds that are different from the training seeds and report the average of the trajectory rewards.

\subsection{Results} \label{sec:results}

\begin{figure}[thb] \centering
    \includegraphics[width=0.3\textwidth]{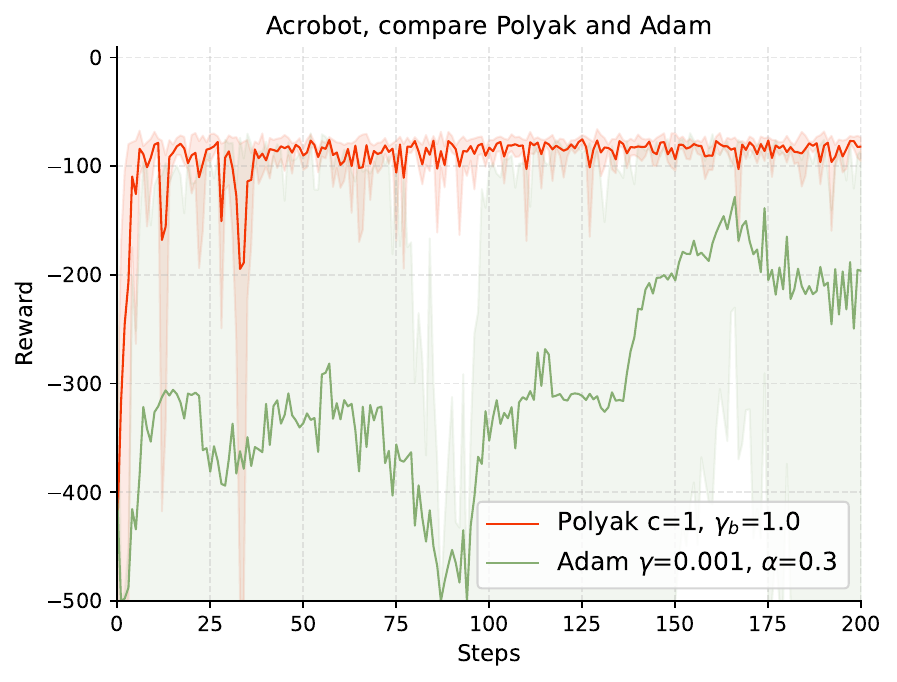}
    \includegraphics[width=0.3\textwidth]{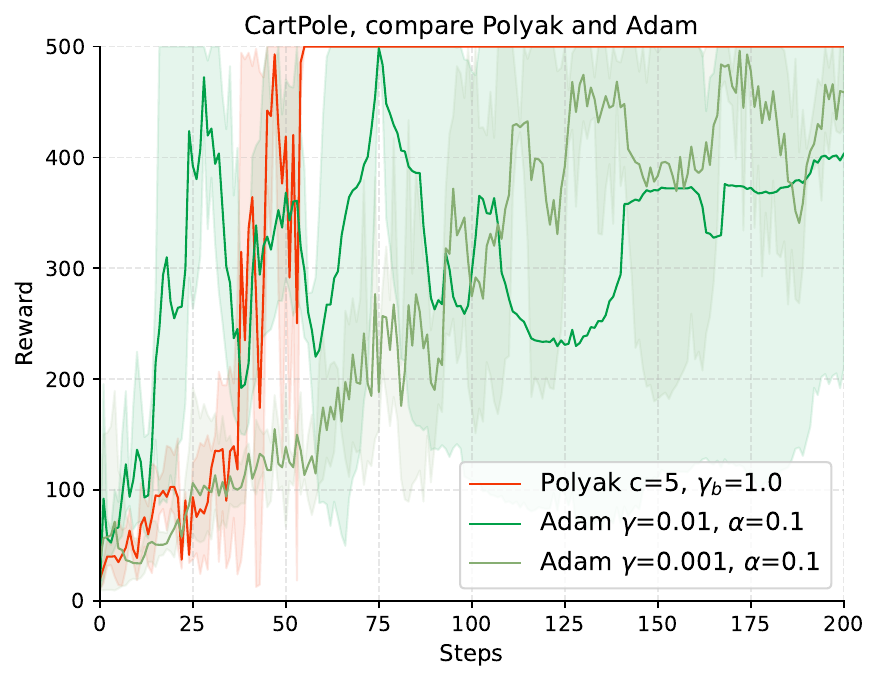}
    \includegraphics[width=0.3\textwidth]{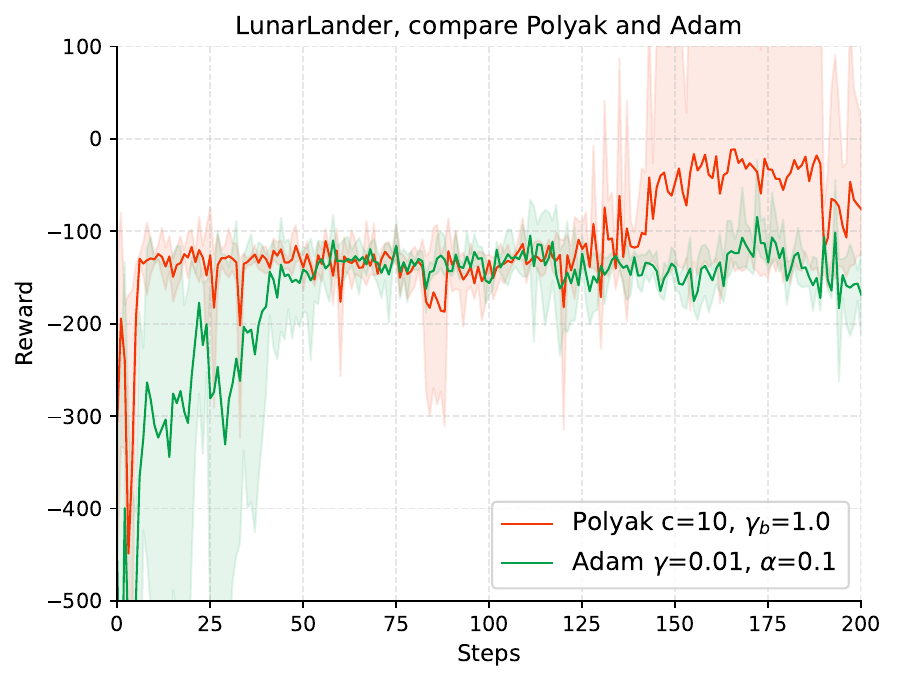}
    \caption{Compare the performance of Adam and the Polyak step-size. We select the best performance of both methods with different hyper-parameters. Polyak step-size converges faster and more stable. The line is the average of three random seeds, and the shade shows the min and max.} 
    \label{fig:compare}
\end{figure}

We investigate multiple hyper-parameter combinations to optimize Adam's performance, while observing consistent stability with the Polyak step-size across diverse hyper-parameters. Due to space constraints, only selected results are depicted in Figure \ref{fig:compare}. Comprehensive results encompassing various hyper-parameter combinations, the trend of step-sizes, and the values of $V^* - V$ are provided in the Appendix. We evaluate both approaches on three Gym environments with discrete action spaces: Acrobot, CartPole, and LunarLander. In Acrobot, where the agent must explore valid trajectories to achieve the target height and improve policy, Adam, while capable of finding such trajectories, exhibits inconsistent improvement. In contrast, Polyak's method consistently leverages successful trajectories to enhance policy. In CartPole, as discussed in Section \ref{sec:intro}, Adam exhibits varying convergence rates with different step-sizes: larger step-sizes lead to fast but unstable convergence, while smaller ones result in slower but more stable convergence. The Polyak step-size, on the other hand, converges optimally and maintains stability. As shown in Figure \ref{fig:compare} and the appendix, the Polyak step-size consistently outperforms Adam in LunarLander.

\begin{figure}[H] \centering
    \includegraphics[width=0.6\textwidth]{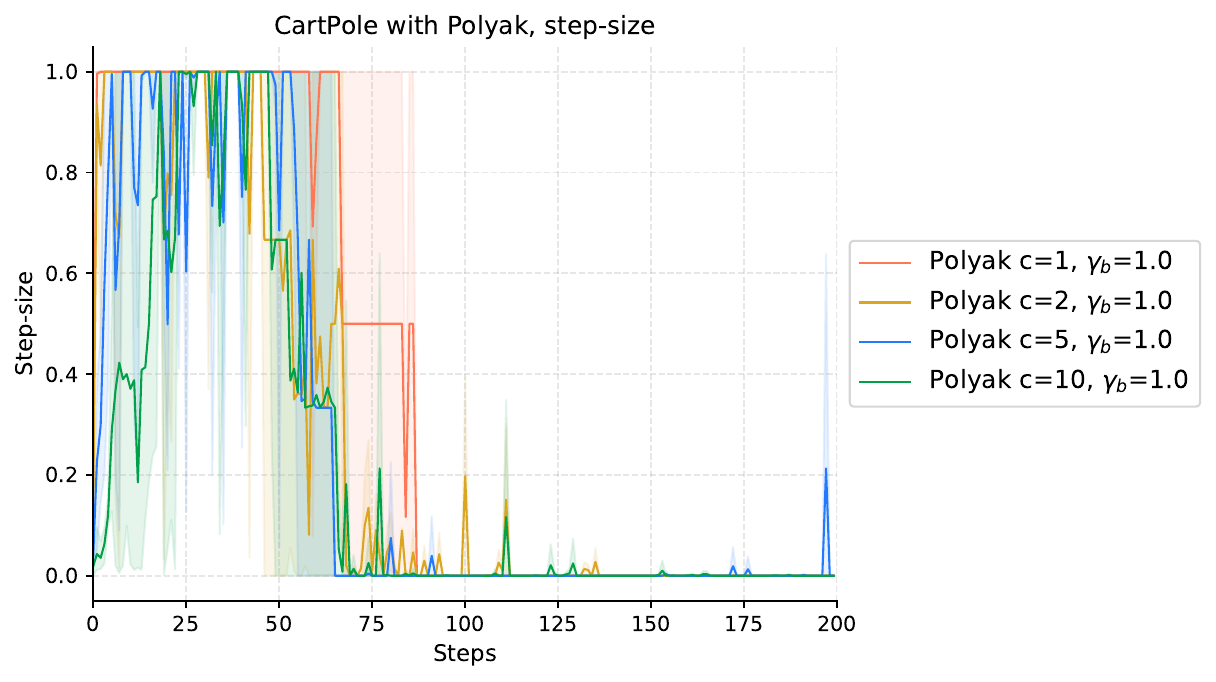}
    \caption{The step-size of the Polyak step-size with twin-model method in CartPole. The step-size decreases when the policy approaches to the optimal policy.} 
    \label{fig:step-size-sample}
\end{figure}

To illustrate the adaptive nature of the Polyak step-size, we include the step-size values from Polyak experiments with $\gamma_b = 1.0$ in Figure \ref{fig:step-size-sample}. Specifically, the line with $c=5$ corresponds to the line in Figure \ref{fig:compare}. It's notable that the step-size rapidly decreases to $0.0$ once the agent discovers a successful policy, ensuring policy stability.

\section{Conclusion}
In this paper, we address the challenge of sensitive step-size tuning in RL by integrating a variant of the Polyak step-size, $\operatorname{SPS}_{\text{max}}$, with policy gradient methods. We investigate concerns related to $V^*$ and explosive updates in softmax policy to ensure the applicability of the Polyak step-size in RL settings. Furthermore, we conduct experiments to evaluate the performance of our proposed method, demonstrating faster convergence and stable policy outcomes.

\bibliography{main}
\bibliographystyle{rlc}

\appendix



\section*{Appendix}

Here, we provide a comprehensive overview of all experiment results.

\subsection*{Performance with Adam}
In this section, we present Adam's performance in the Acrobot, CartPole, and LunarLander environments. We compare different step-sizes $\operatorname{lr}$ and also include exploration coefficient $\alpha$ as mentioned in Section \ref{sec:stoc_issue}. As outlined in Section \ref{sec:results}, we observe that relatively large step-sizes lead to fast convergence but unstable results, while small step-sizes result in slower convergence but slightly greater stability.

\begin{figure}[H] \centering
    \includegraphics[width=0.45\textwidth]{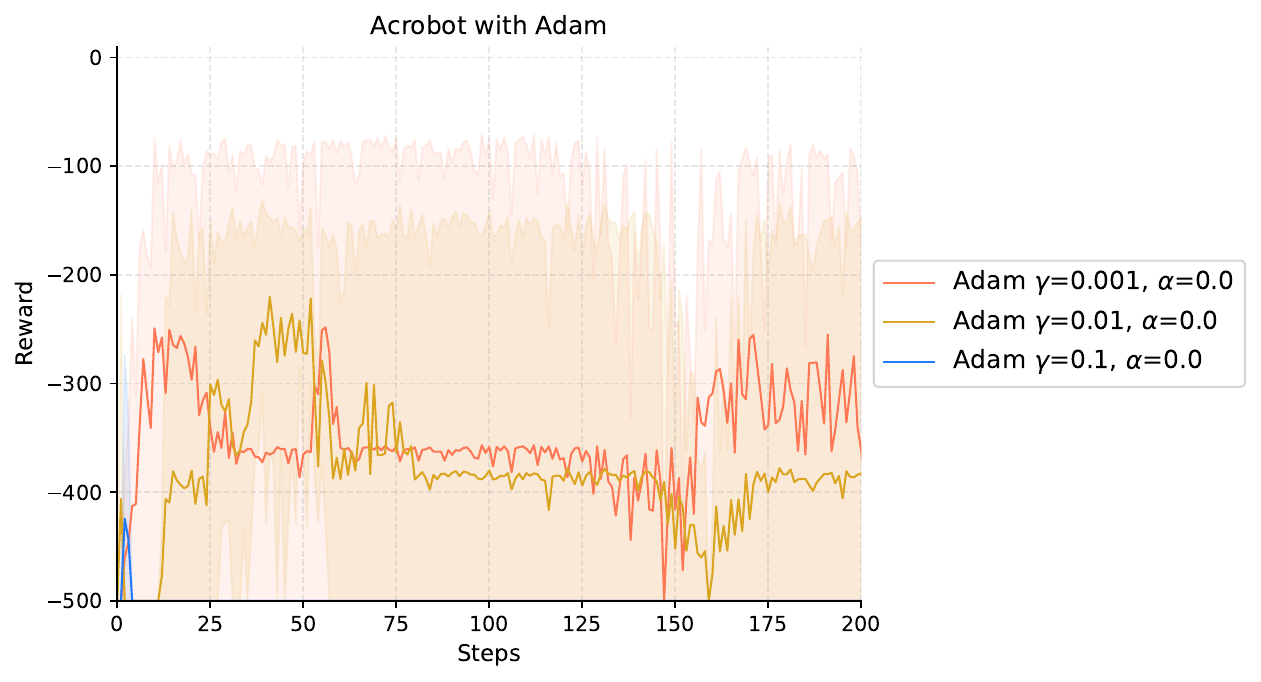}
    \includegraphics[width=0.45\textwidth]{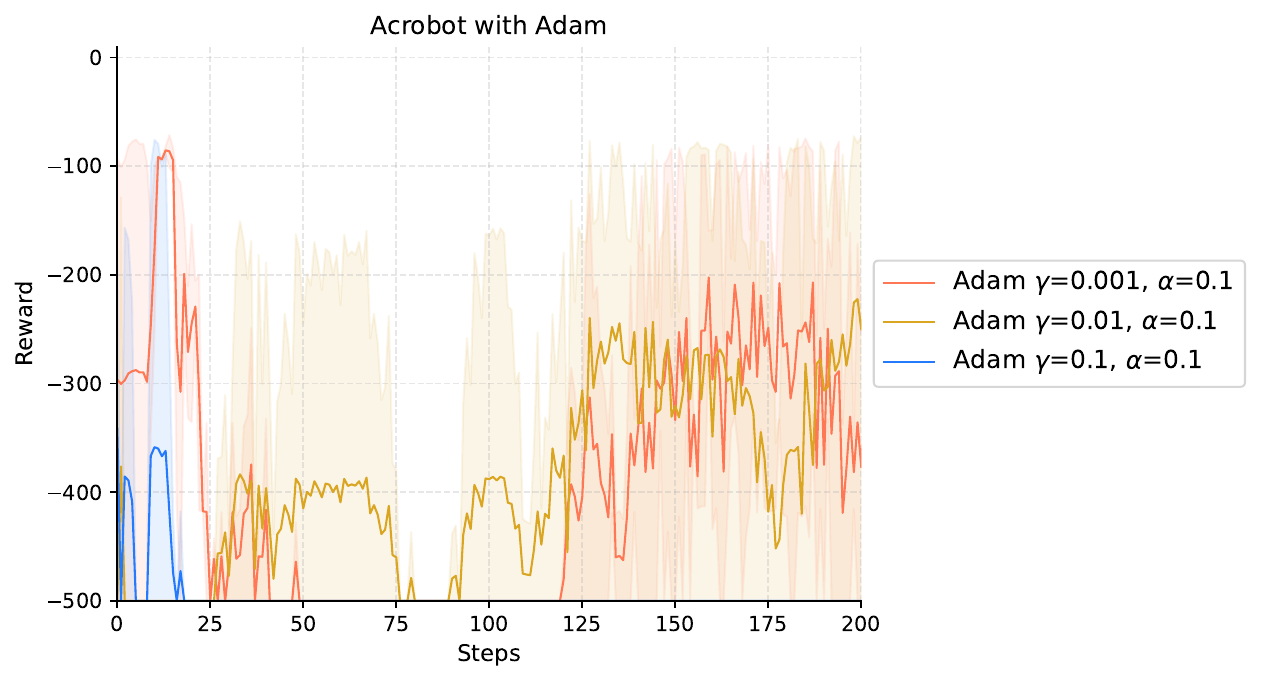}
    \\
    \includegraphics[width=0.45\textwidth]{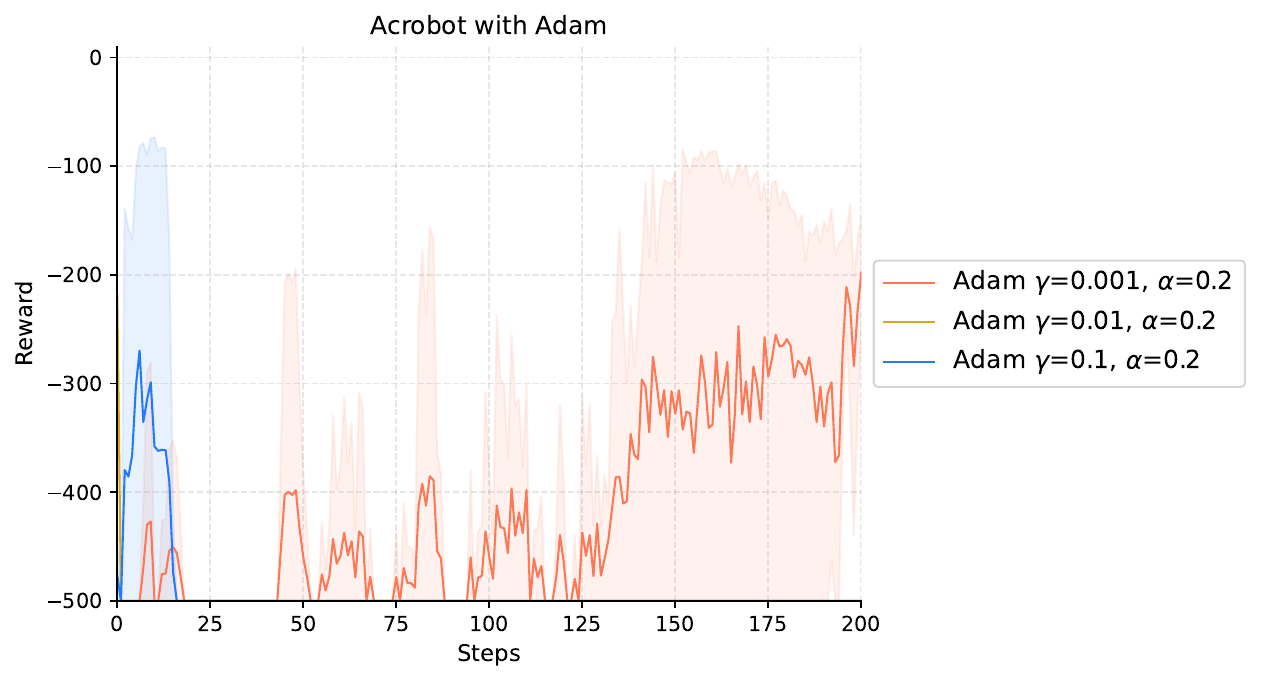}
    \includegraphics[width=0.45\textwidth]{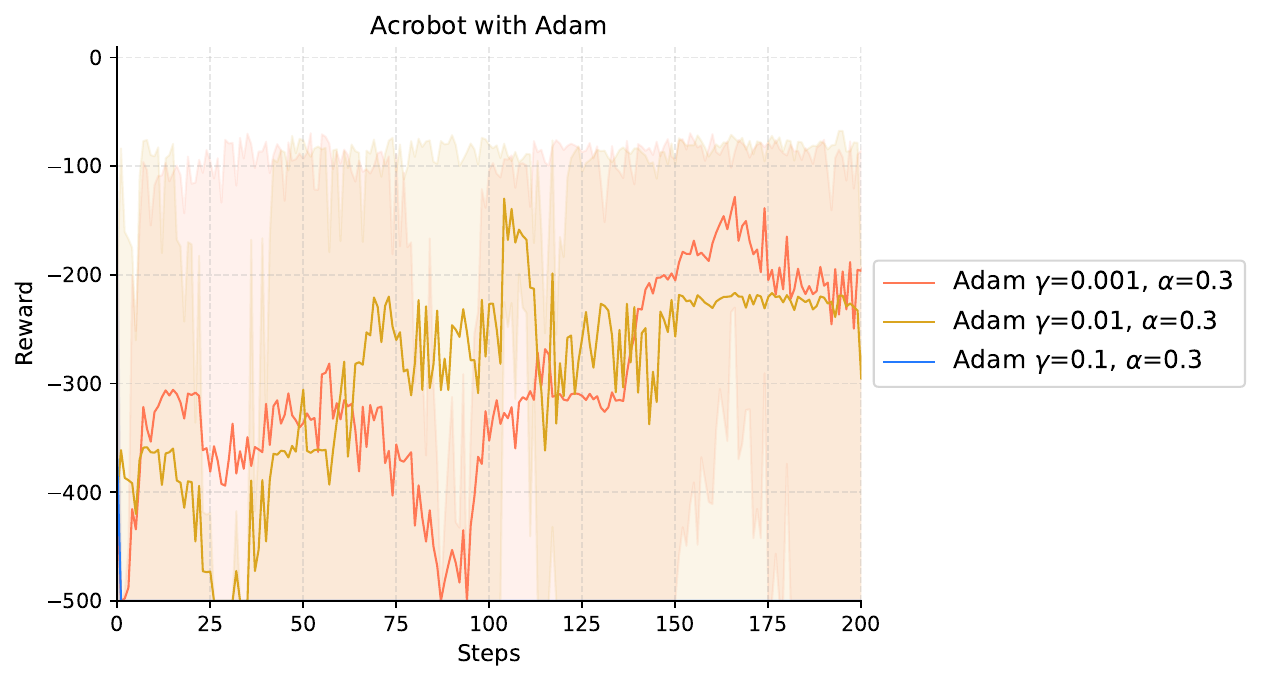}
    \caption{Performance of Adam in Acrobot.}
\end{figure}

\begin{figure}[H] \centering
    \includegraphics[width=0.45\textwidth]{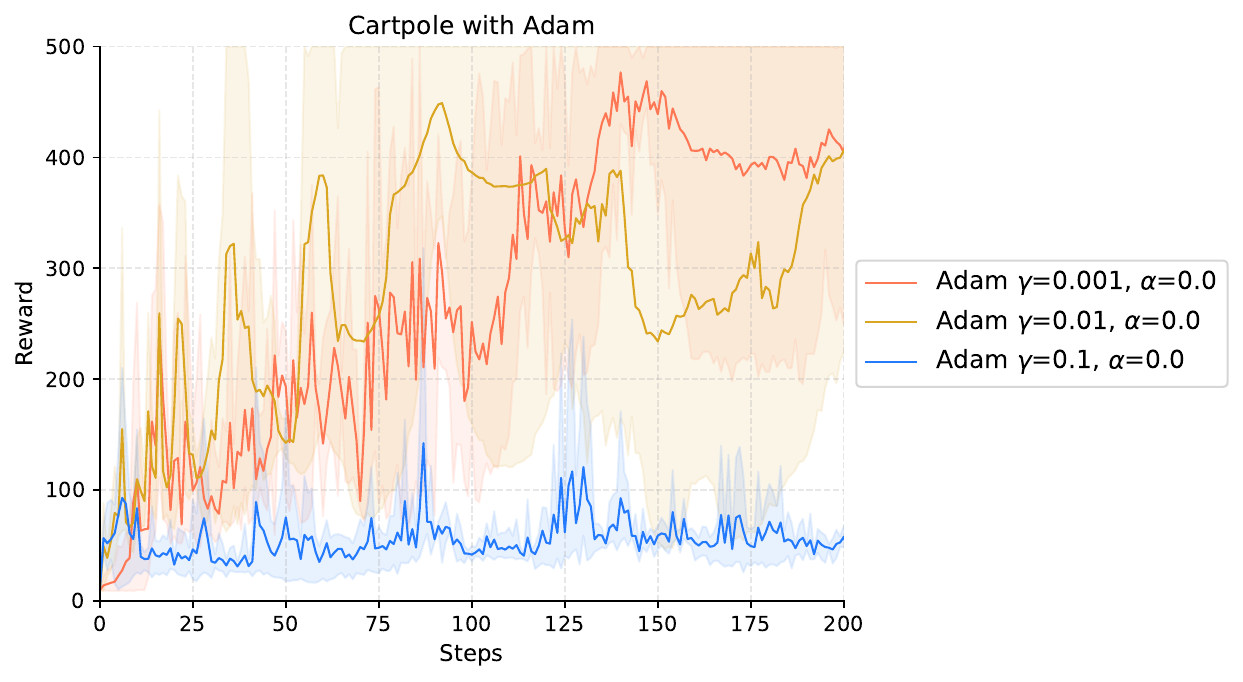}
    \includegraphics[width=0.45\textwidth]{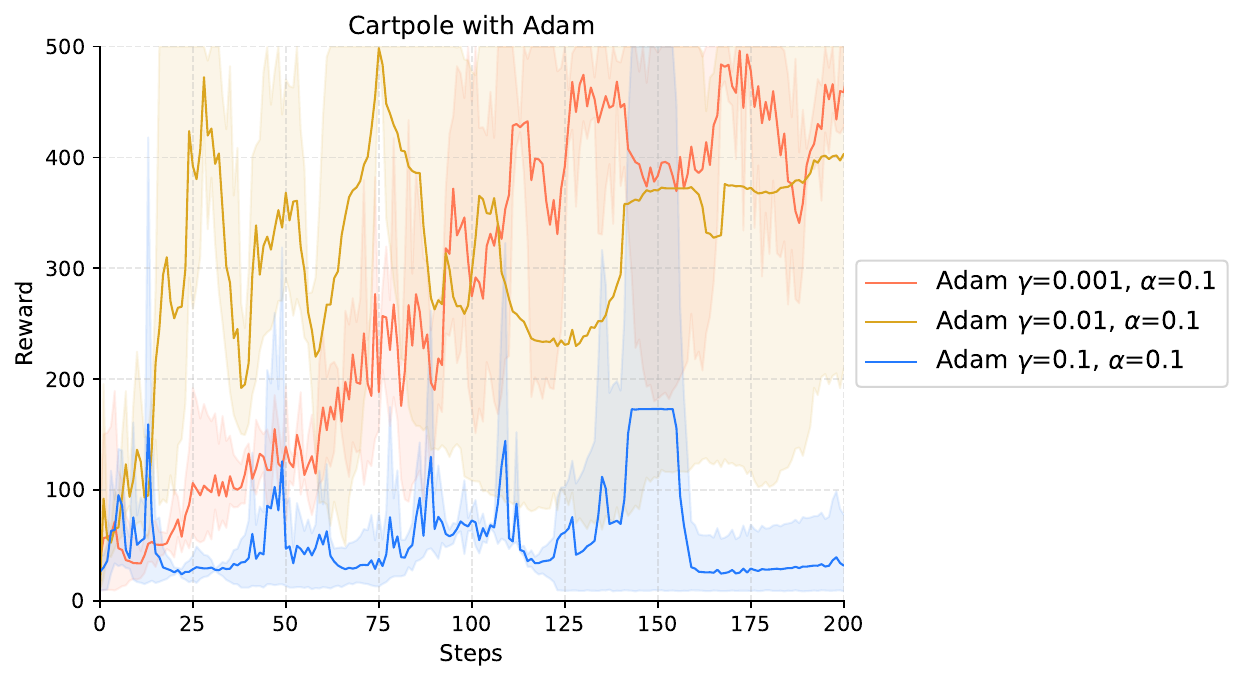}
    \\
    \includegraphics[width=0.45\textwidth]{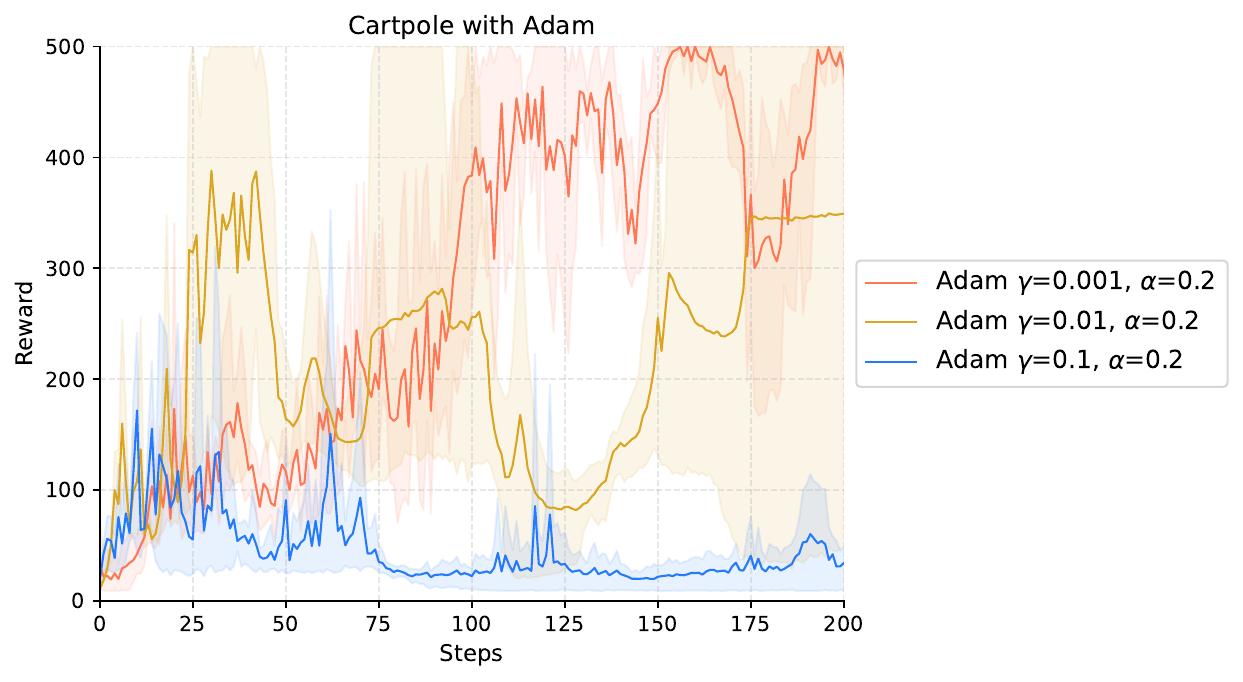}
    \includegraphics[width=0.45\textwidth]{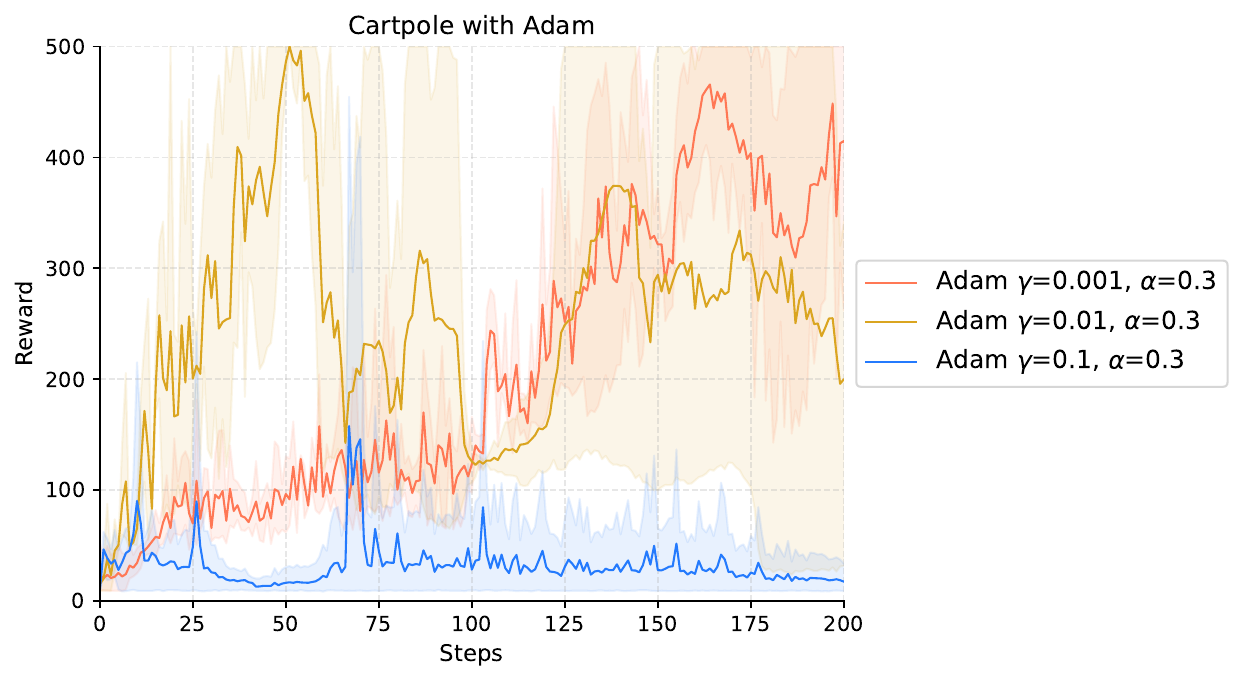}
    \caption{Performance of Adam in CartPole.}
\end{figure}

\begin{figure}[H] \centering
    \includegraphics[width=0.45\textwidth]{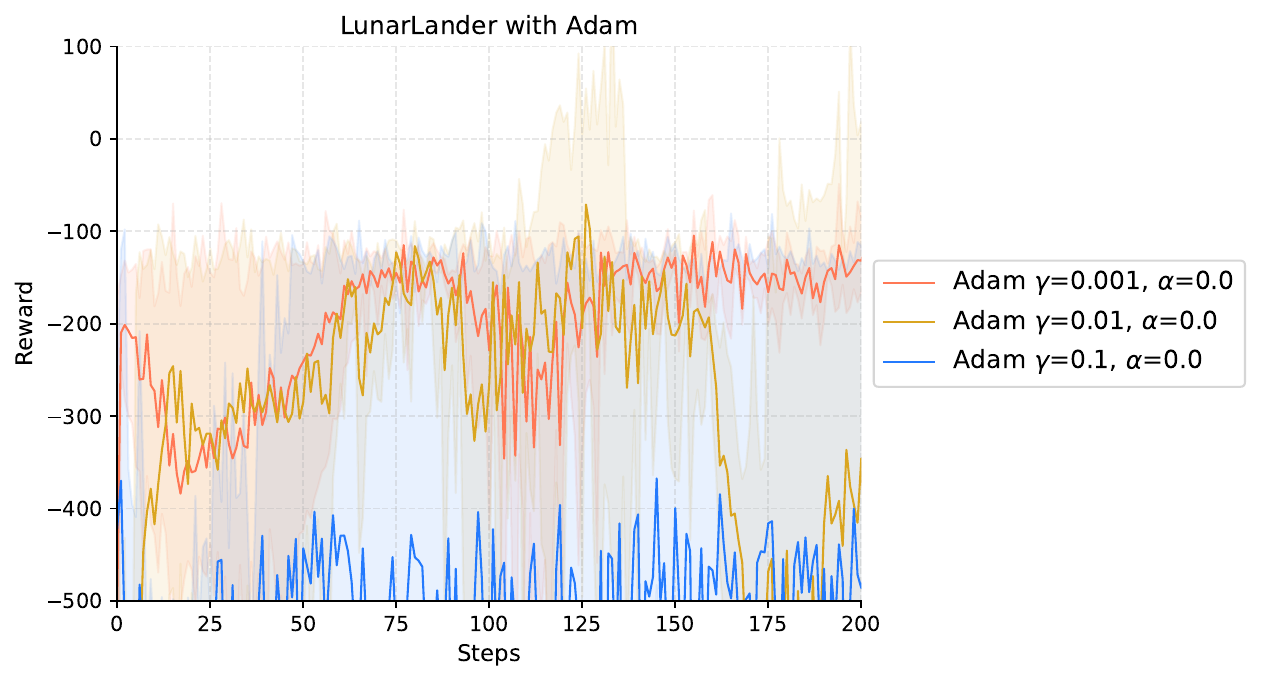}
    \includegraphics[width=0.45\textwidth]{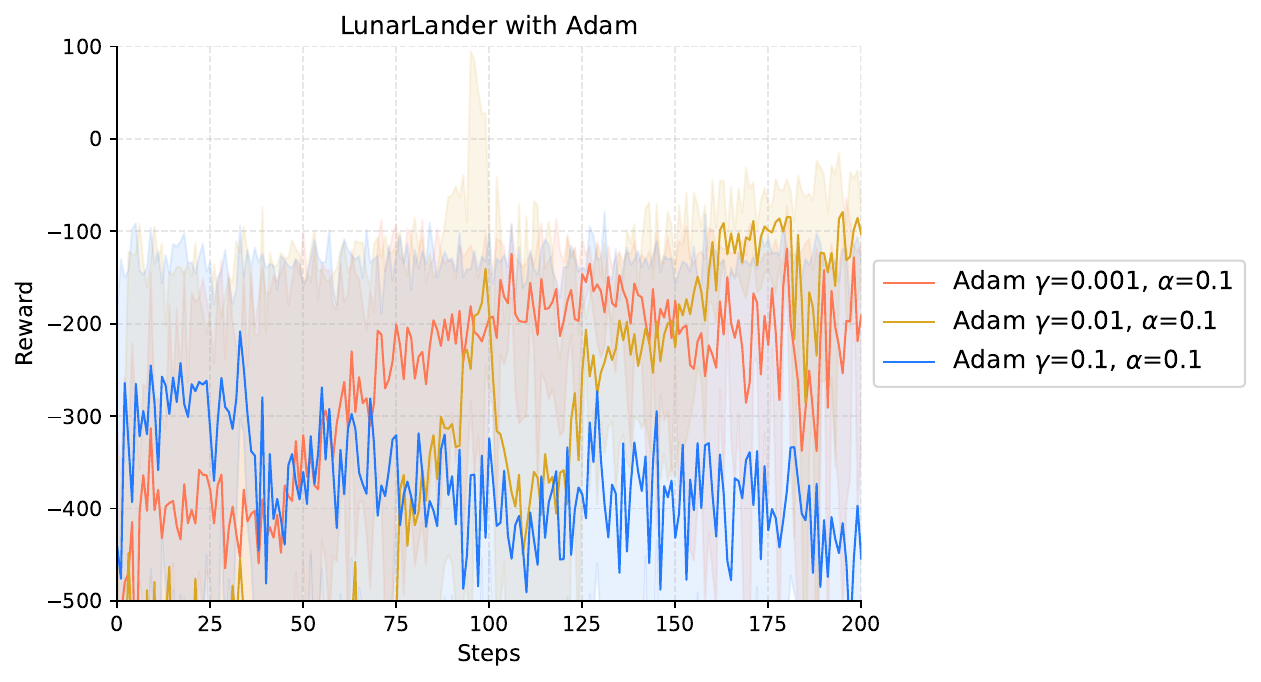}
    \\
    \includegraphics[width=0.45\textwidth]{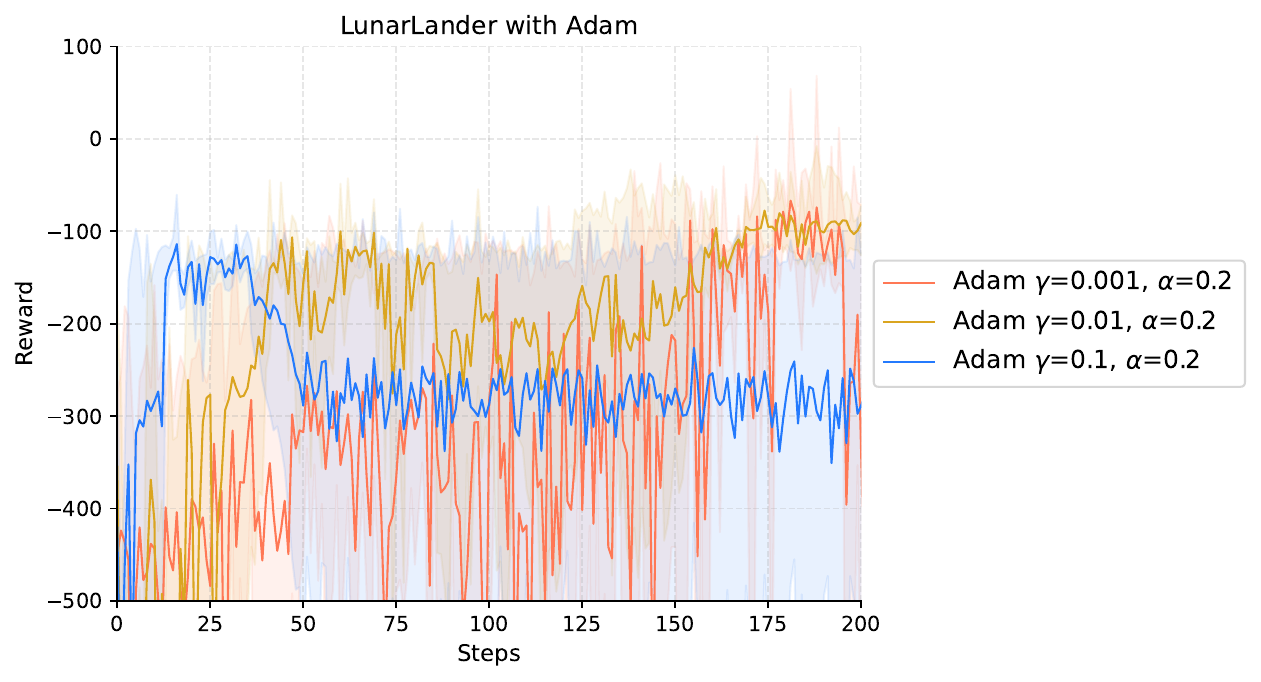}
    \includegraphics[width=0.45\textwidth]{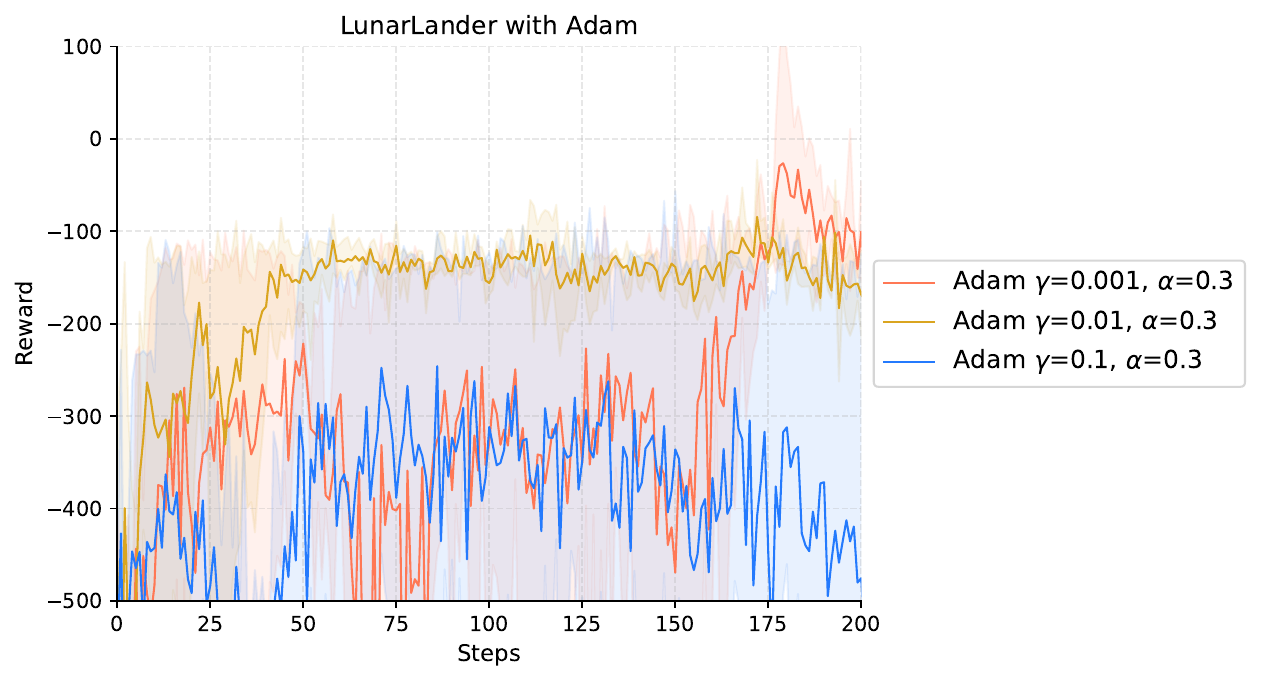}
    \caption{Performance of Adam in LunarLander.}
\end{figure}

\subsection*{Performance with Polyak}

Here, we showcase the performance of the Polyak method with varying values of $c$ and $\gamma_b$ in the environment Acrobot, CartPole and LunarLander. Our method exhibits robustness across different hyper-parameter combinations.

\begin{figure}[H] \centering
    \includegraphics[width=0.45\textwidth]{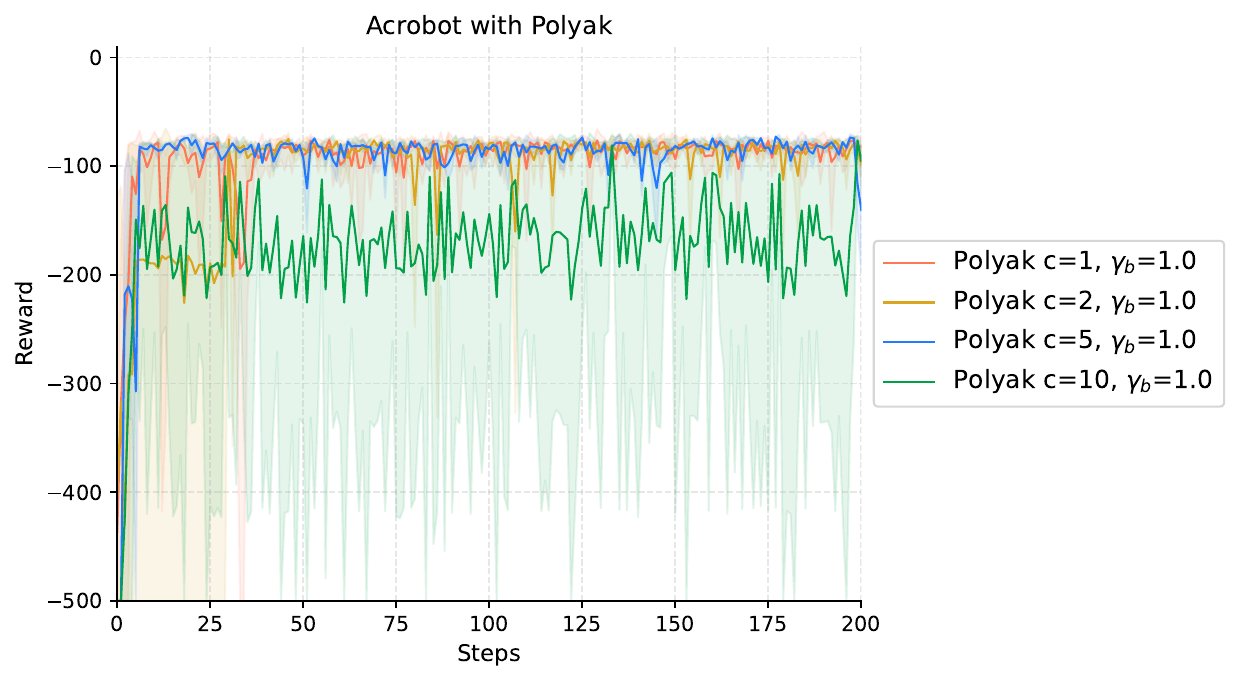}
    \includegraphics[width=0.45\textwidth]{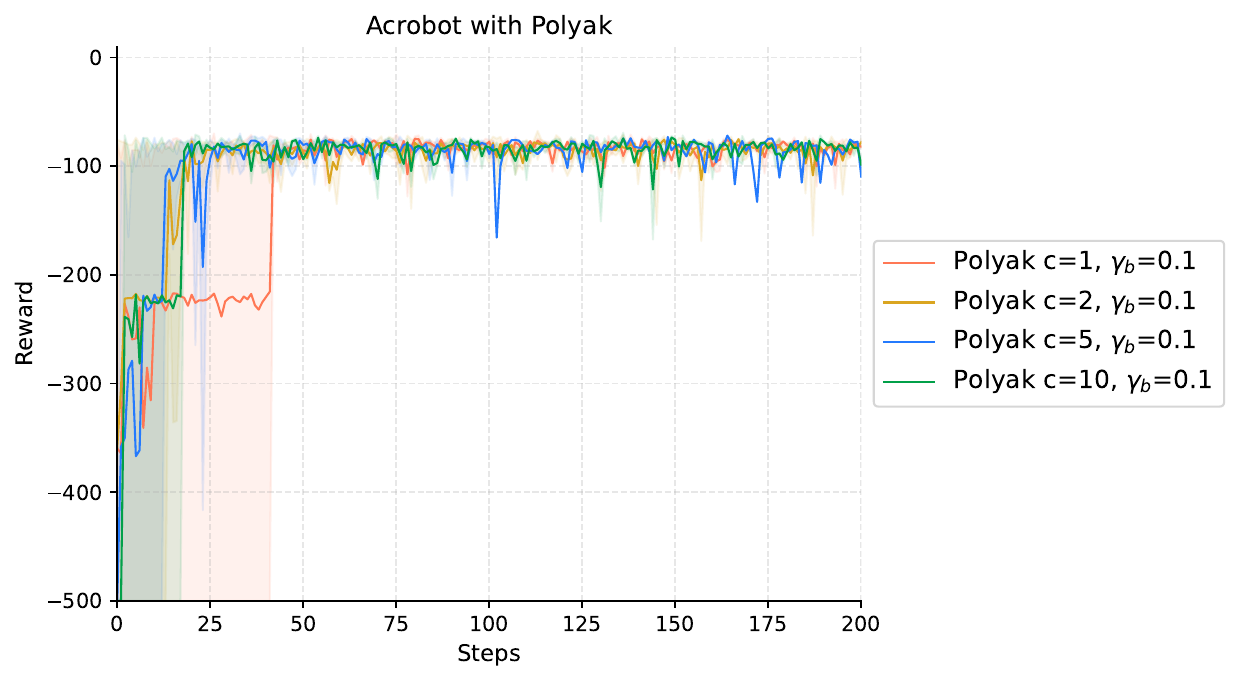}
    \\
    \includegraphics[width=0.45\textwidth]{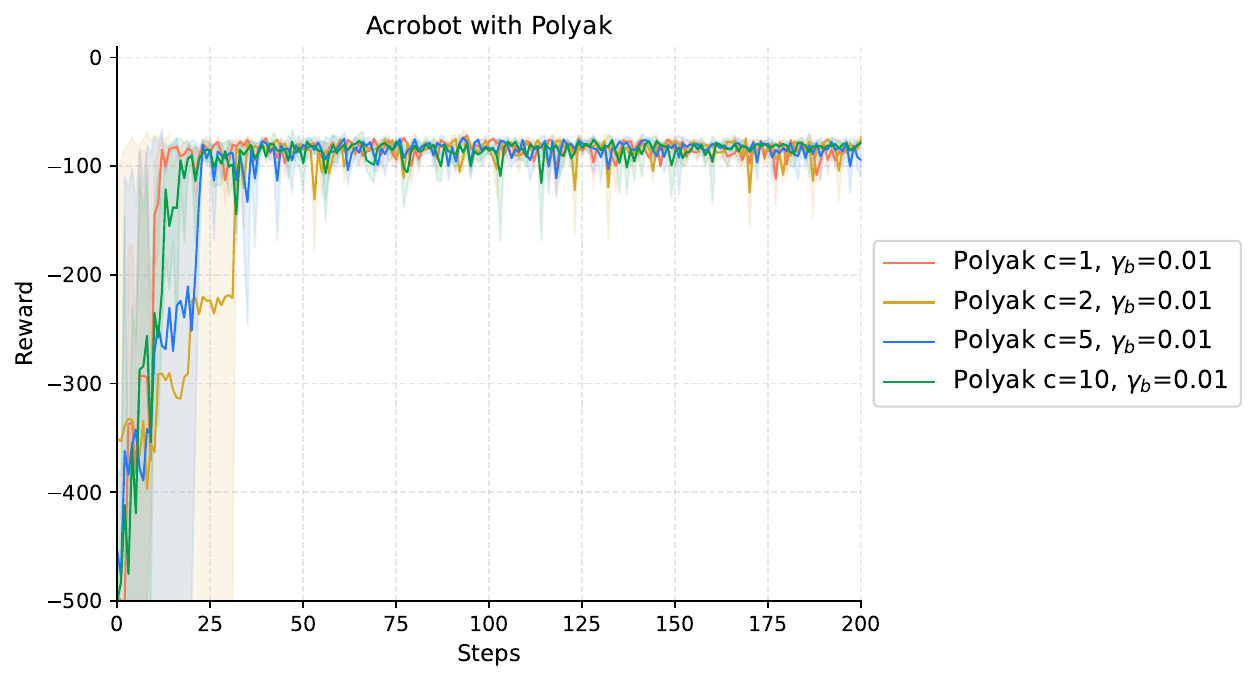}
    \caption{Performance of Polyak in Acrobot.}
\end{figure}

\begin{figure}[H] \centering
    \includegraphics[width=0.45\textwidth]{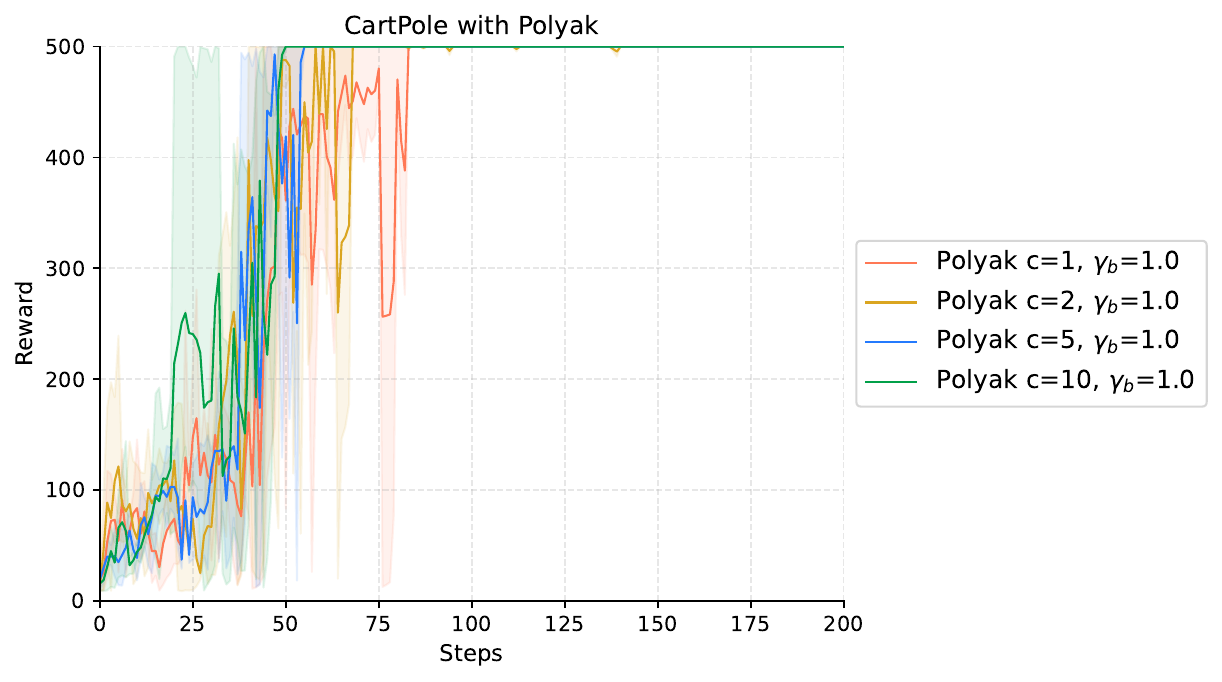}
    \includegraphics[width=0.45\textwidth]{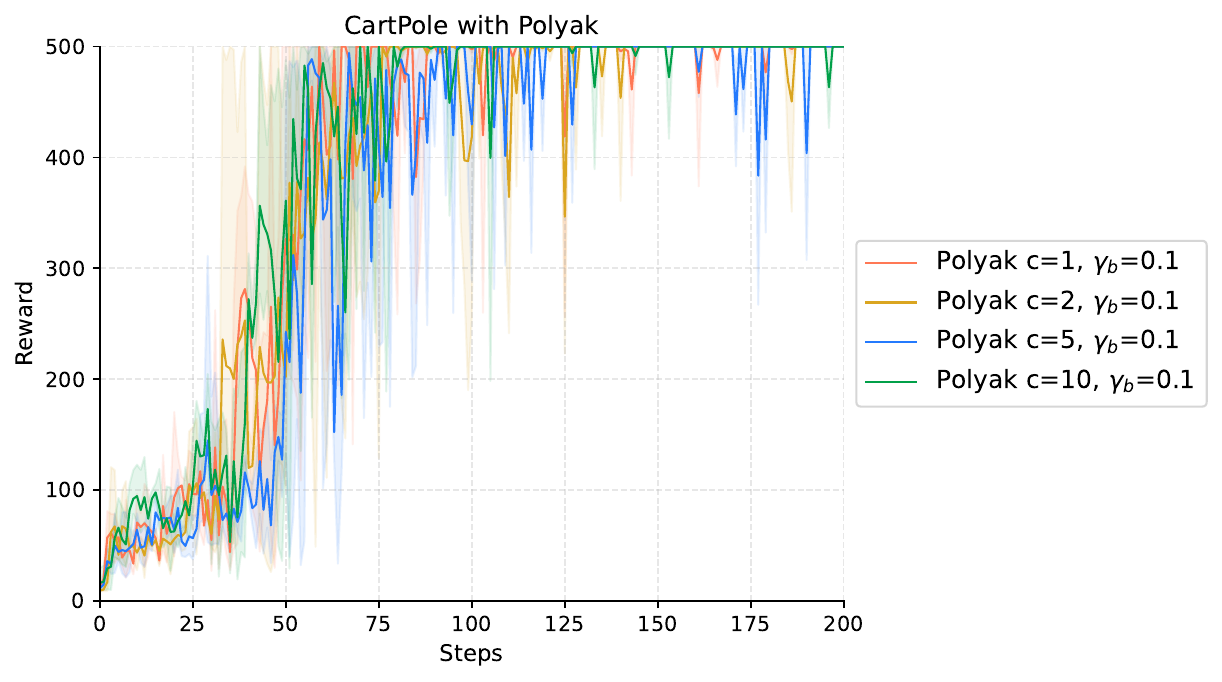}
    \\
    \includegraphics[width=0.45\textwidth]{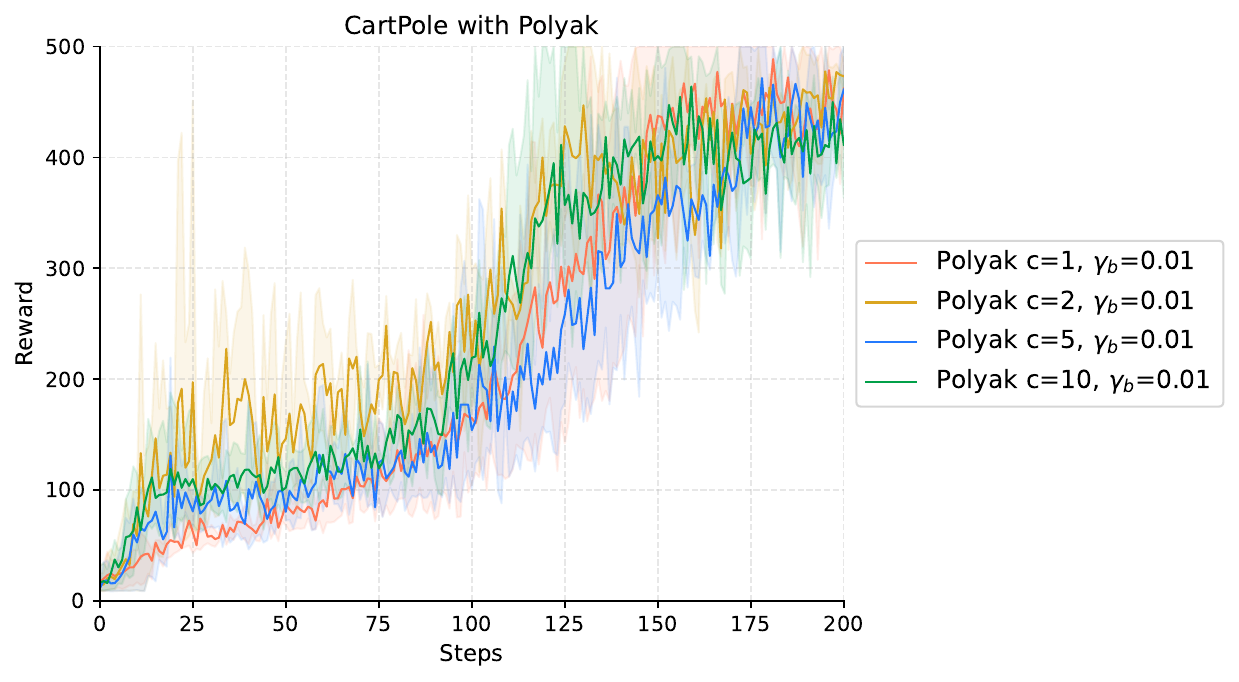}
    \caption{Performance of Polyak in CartPole.}
\end{figure}

\begin{figure}[H] \centering
    \includegraphics[width=0.45\textwidth]{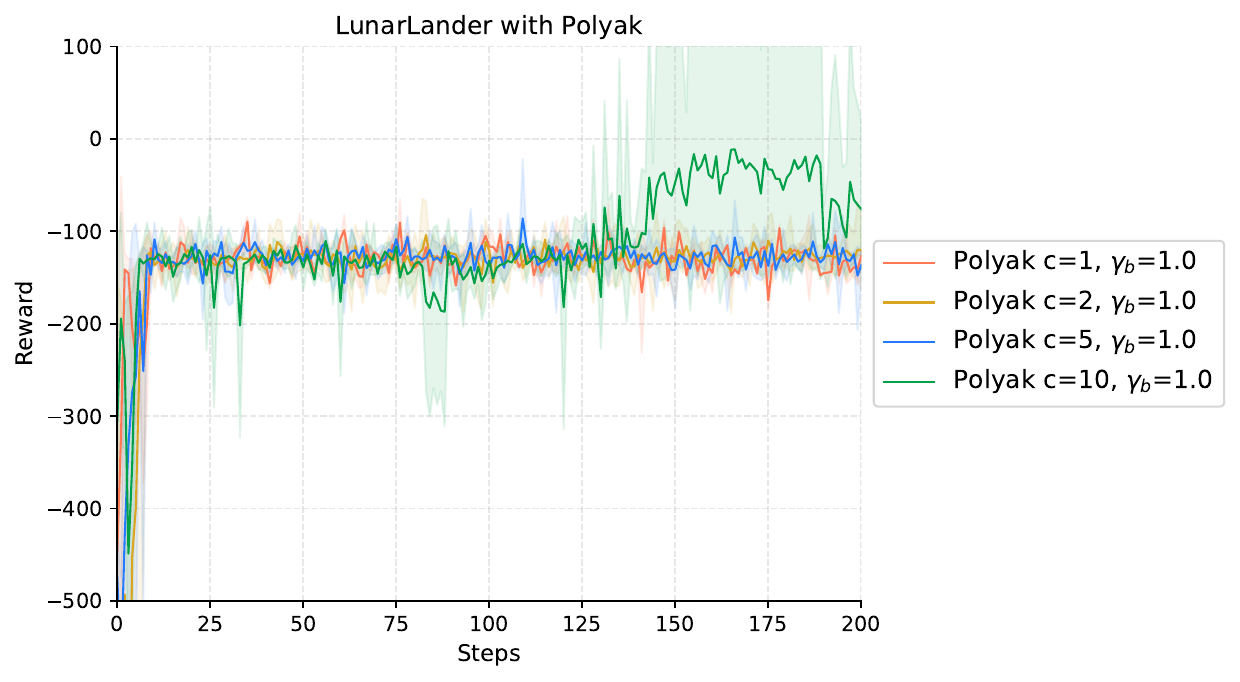}
    \includegraphics[width=0.45\textwidth]{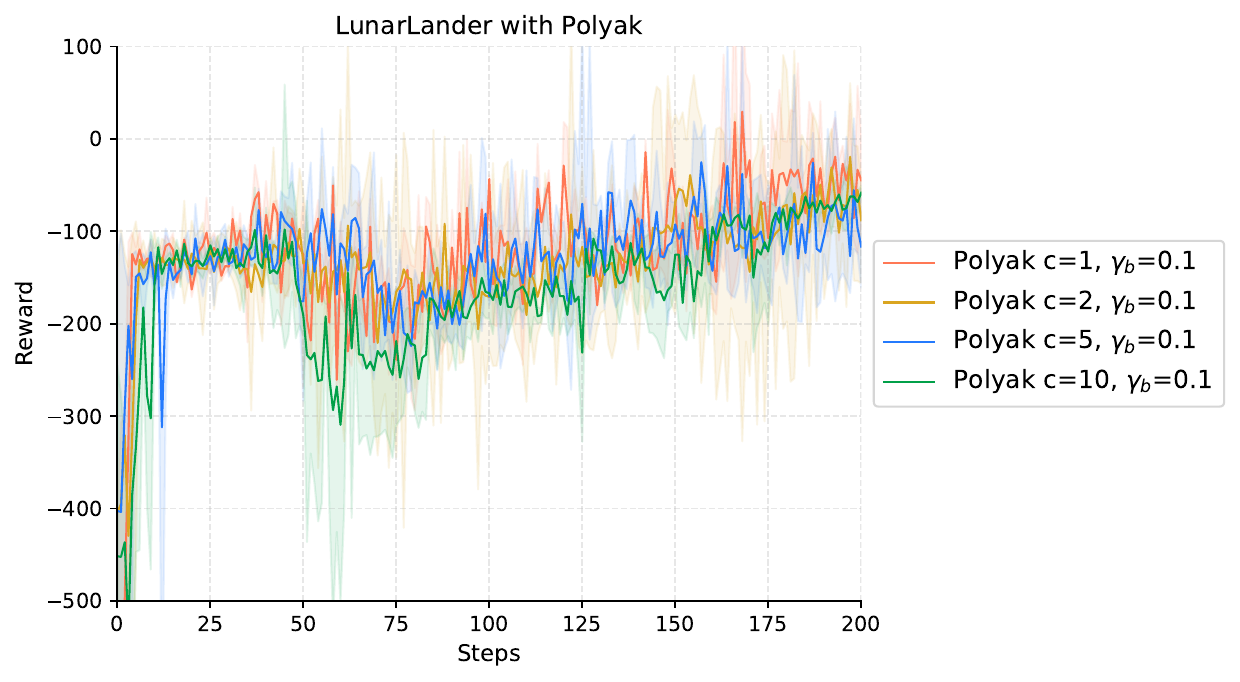}
    \\
    \includegraphics[width=0.45\textwidth]{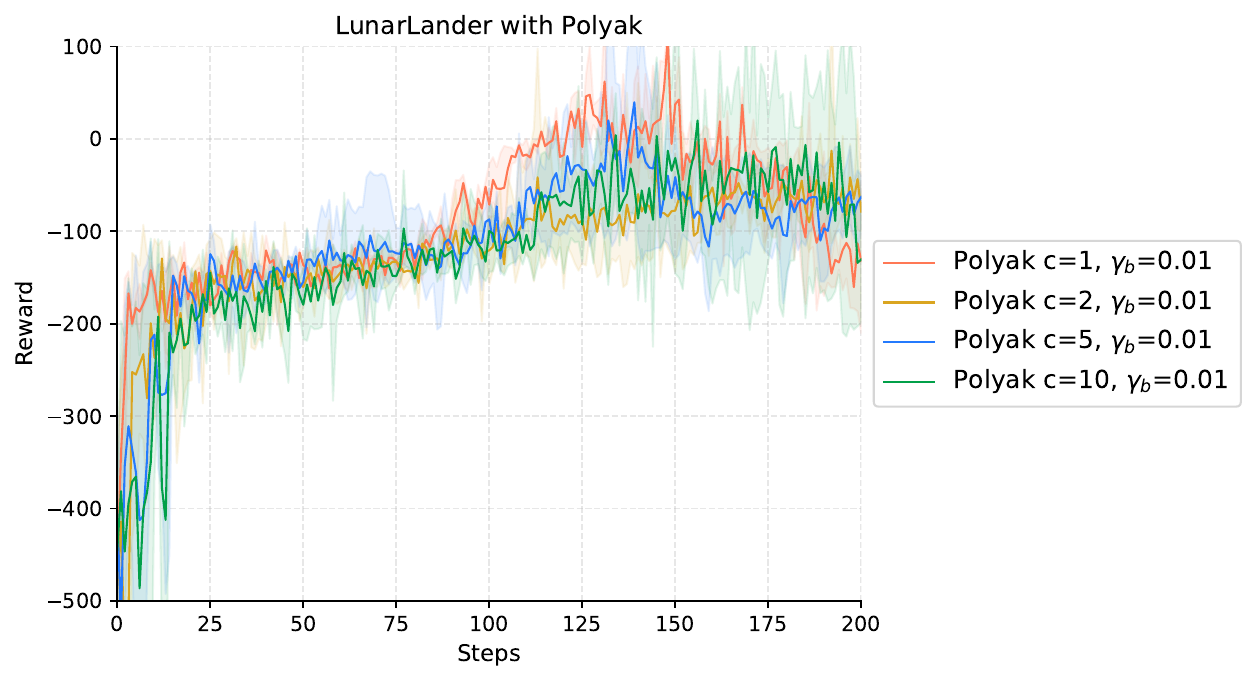}
    \caption{Performance of Polyak in LunarLander.}
\end{figure}

\subsection*{Step-size in Polyak method}

Here, we display the step sizes corresponding to the Polyak experiments. It is evident that our method automatically adjusts the step size.

\begin{figure}[H] \centering
    \includegraphics[width=0.45\textwidth]{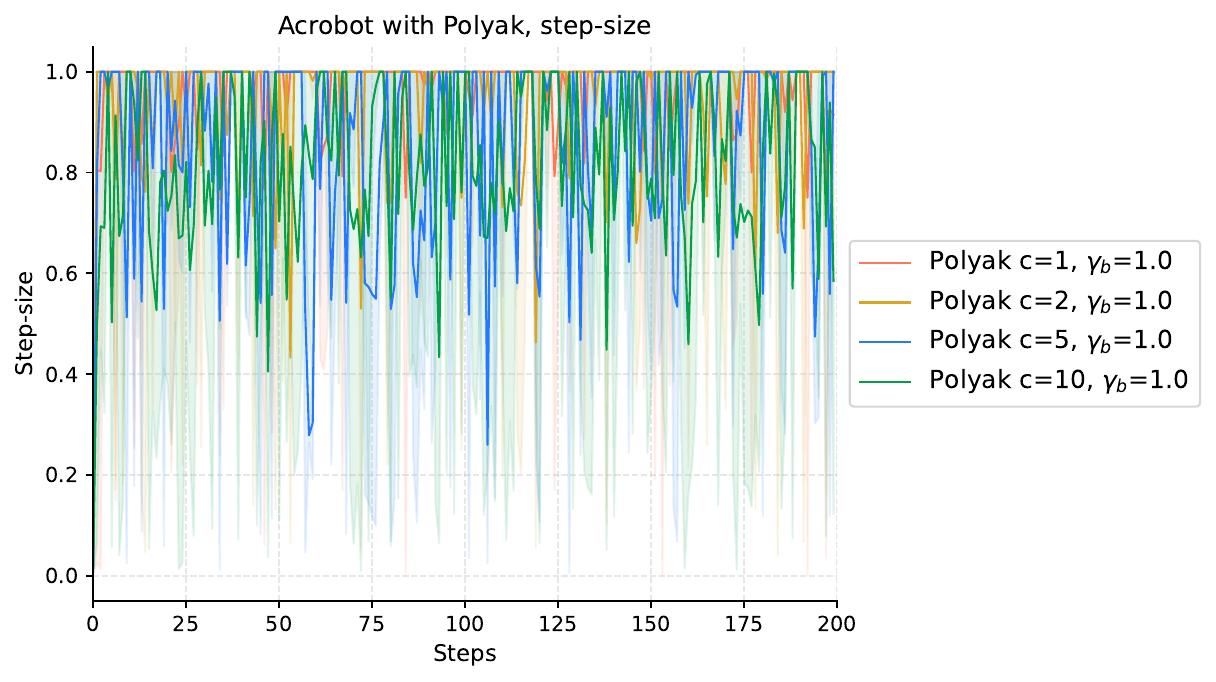}
    \includegraphics[width=0.45\textwidth]{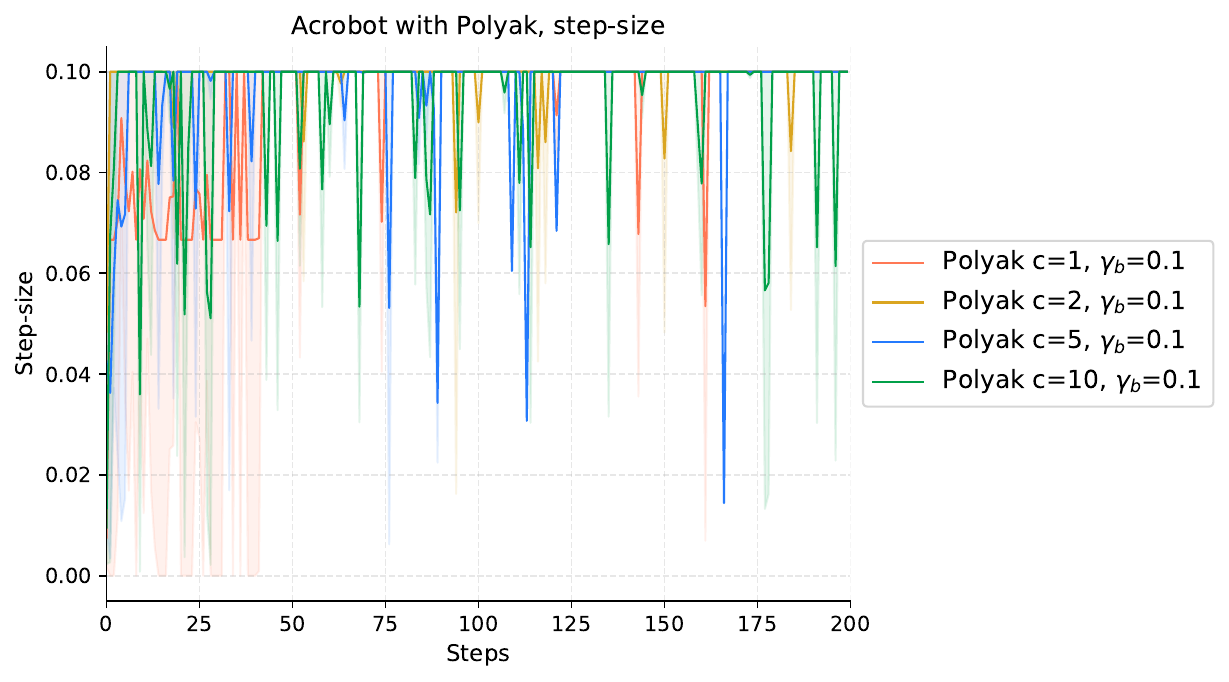}
    \\
    \includegraphics[width=0.45\textwidth]{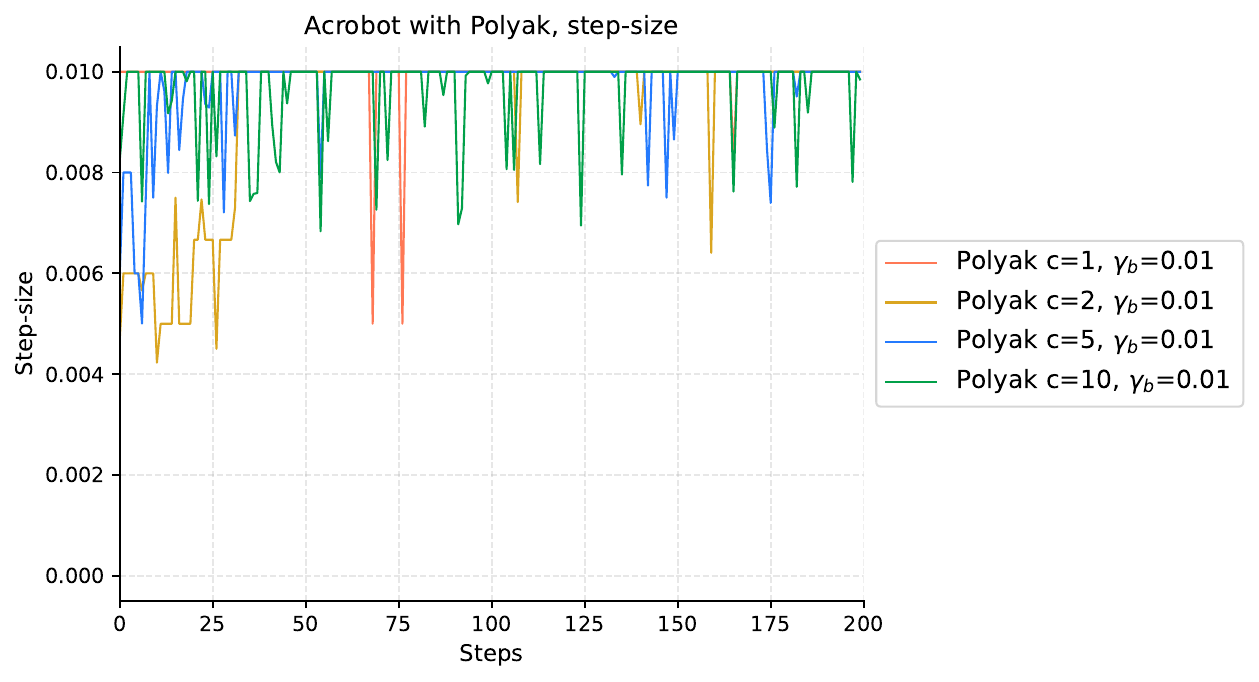}
    \caption{Step-size of Polyak method in Acrobot.}
\end{figure}

\begin{figure}[H] \centering
    \includegraphics[width=0.45\textwidth]{cartpole_polyak_stepsize_gamma1.pdf}
    \includegraphics[width=0.45\textwidth]{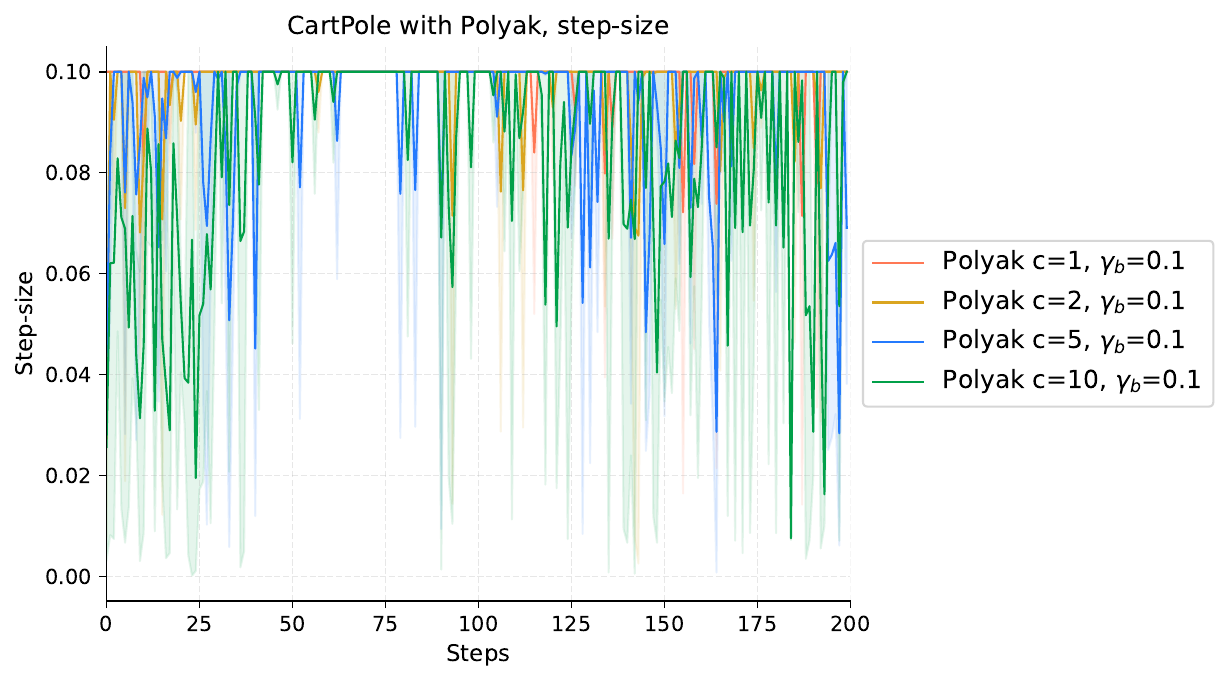}
    \\
    \includegraphics[width=0.45\textwidth]{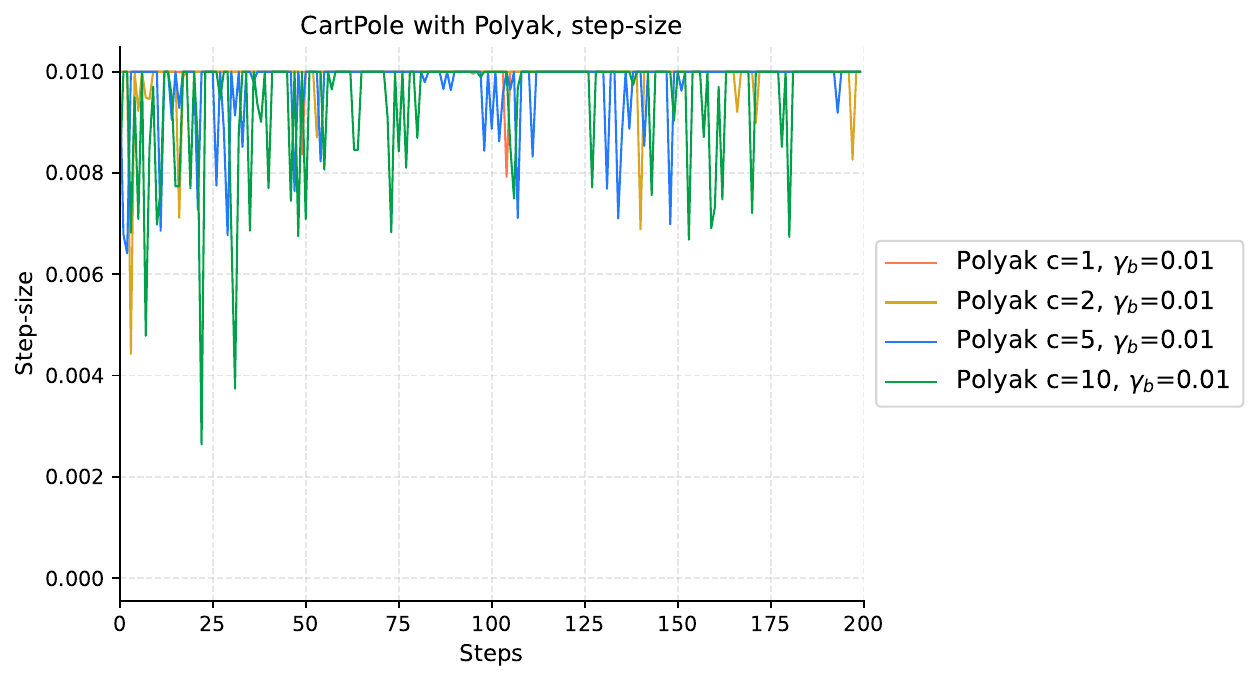}
    \caption{step-size of Polyak method in CartPole.}
\end{figure}

\begin{figure}[H] \centering
    \includegraphics[width=0.45\textwidth]{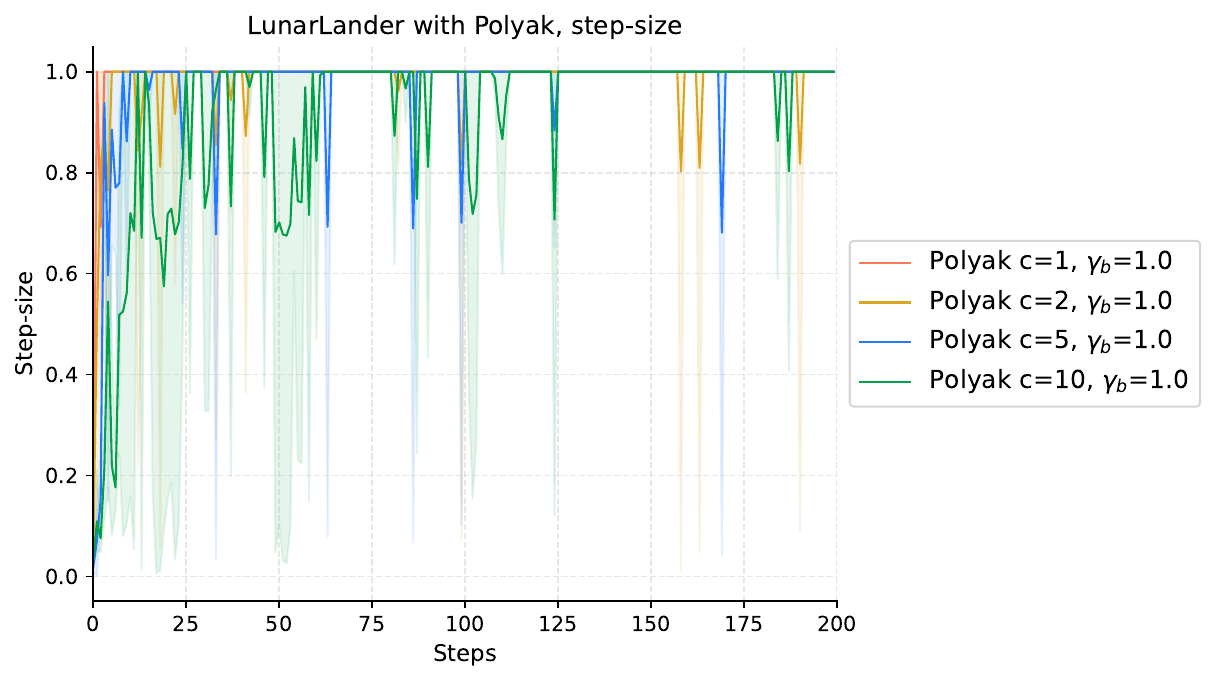}
    \includegraphics[width=0.45\textwidth]{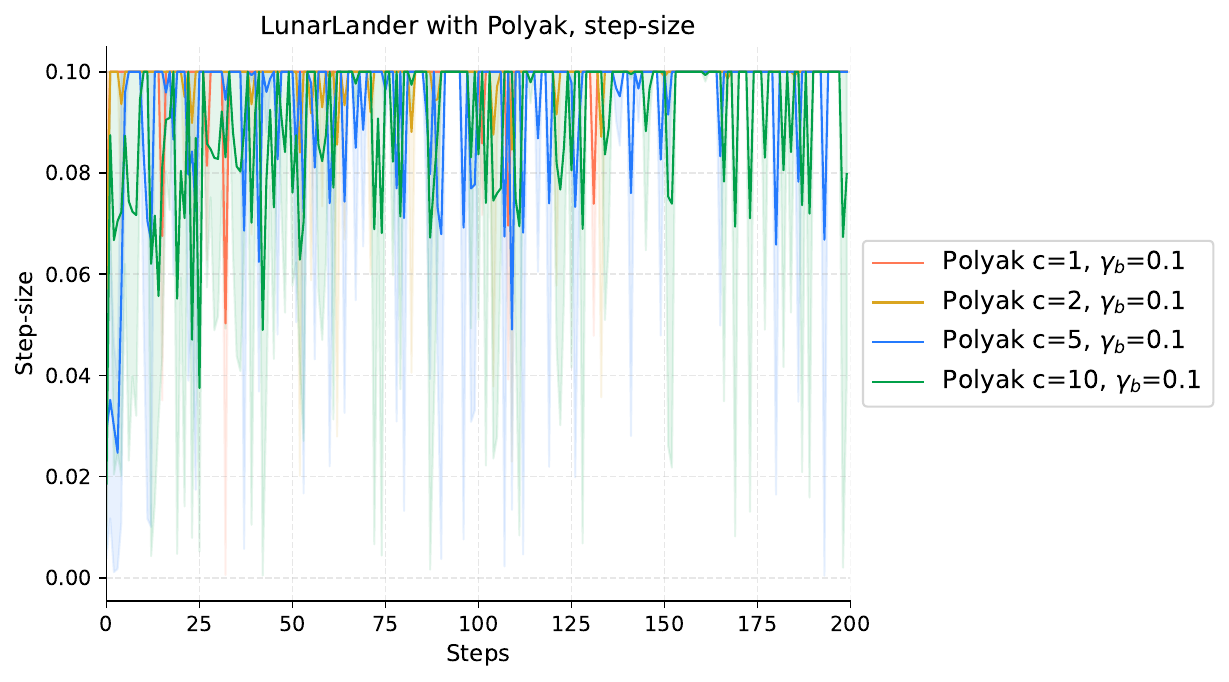}
    \\
    \includegraphics[width=0.45\textwidth]{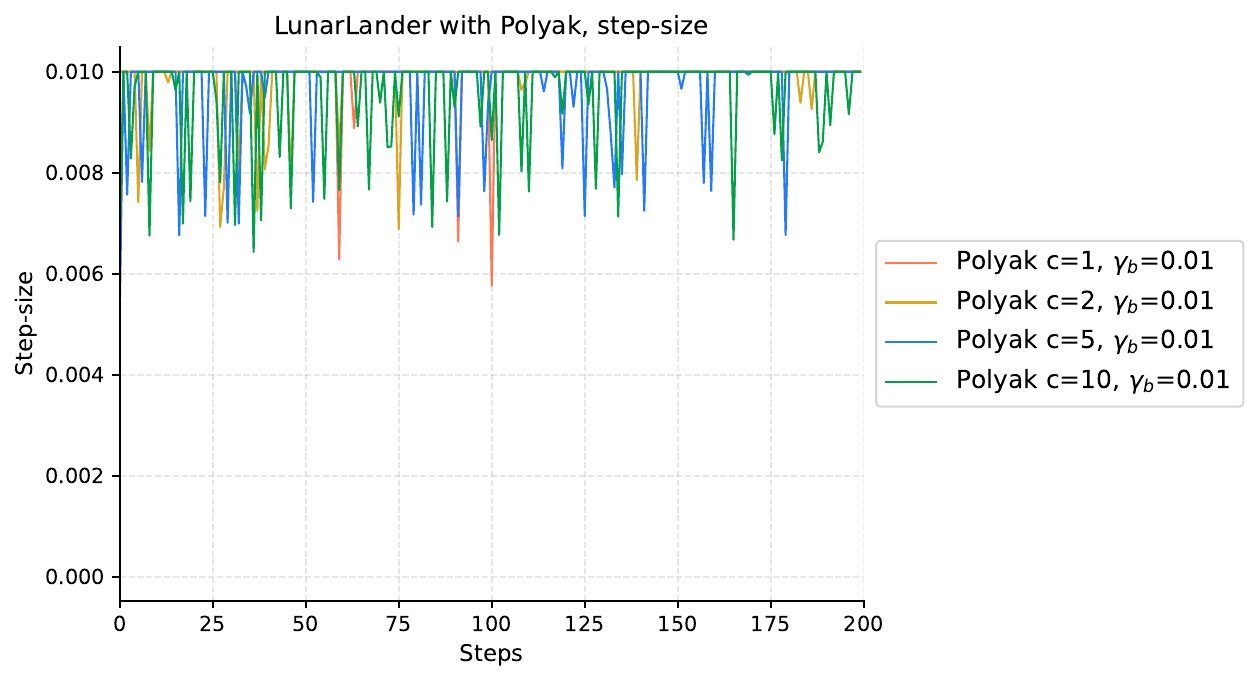}
    \caption{step-size of Polyak method in LunarLander.}
\end{figure}

\subsection*{Value of $V^*-V$ in Polyak method}

Here, we present the difference between the value functions $V^*$ and $V$, where $V^*$ represents the larger value function and $V$ represents the smaller one between two models. We illustrate how both models are updated simultaneously.

\begin{figure}[H] \centering
    \includegraphics[width=0.45\textwidth]{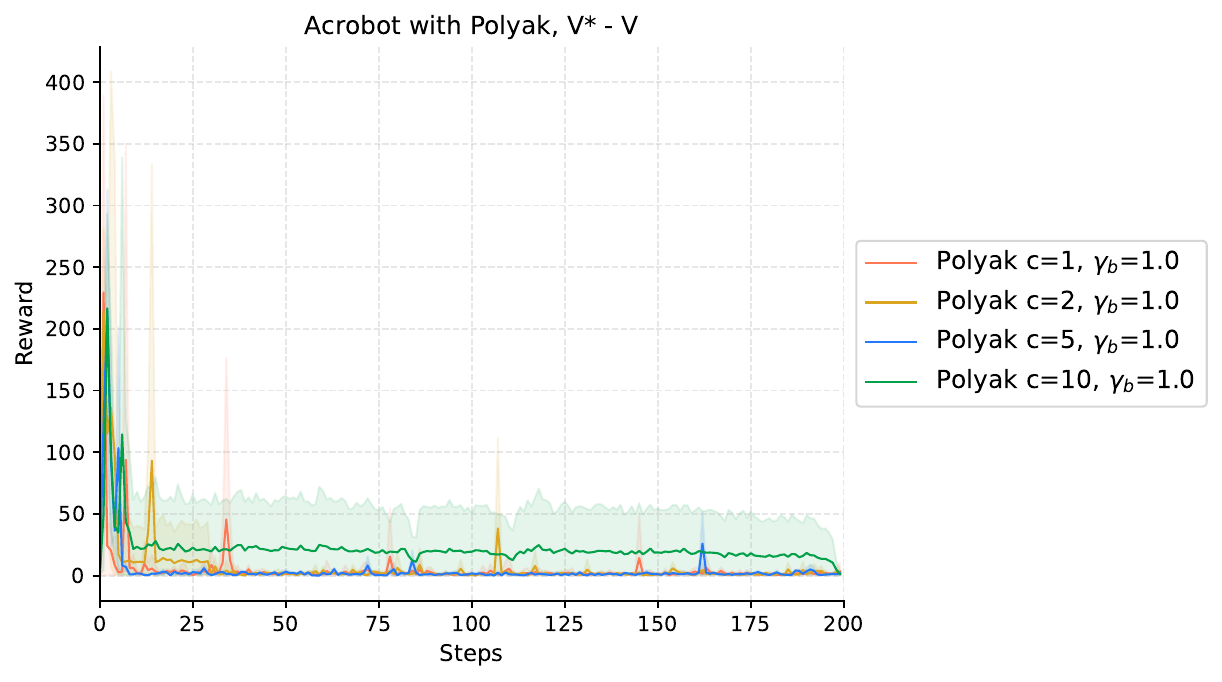}
    \includegraphics[width=0.45\textwidth]{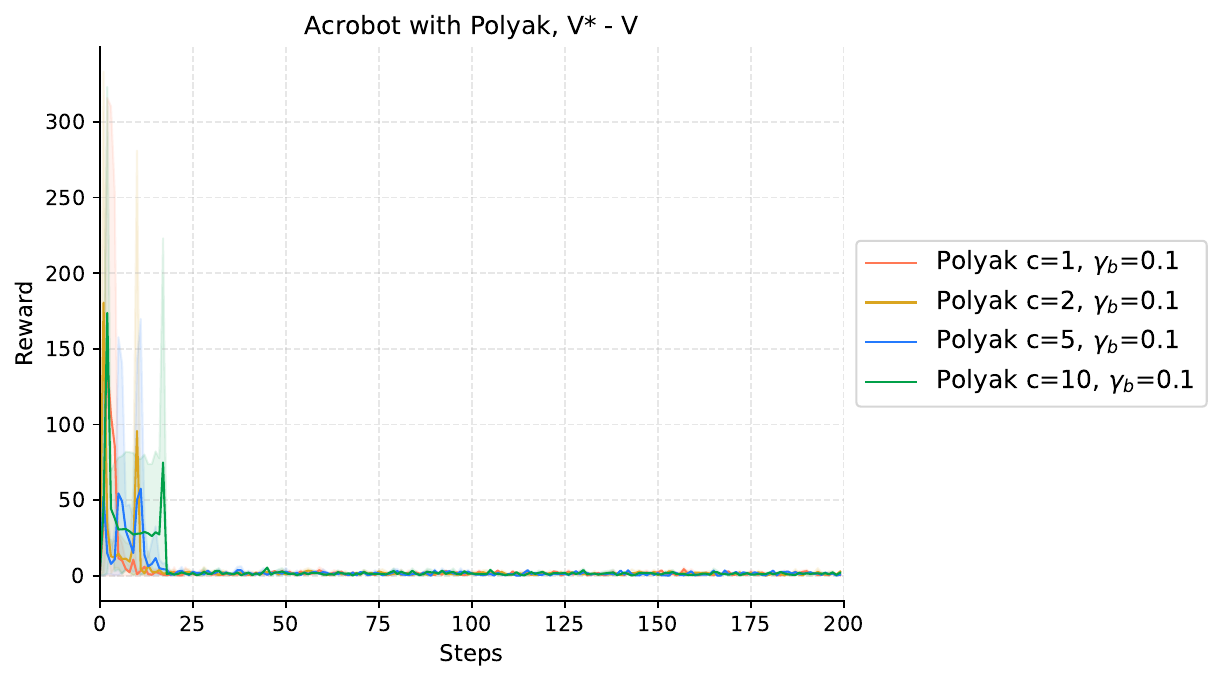}
    \\
    \includegraphics[width=0.45\textwidth]{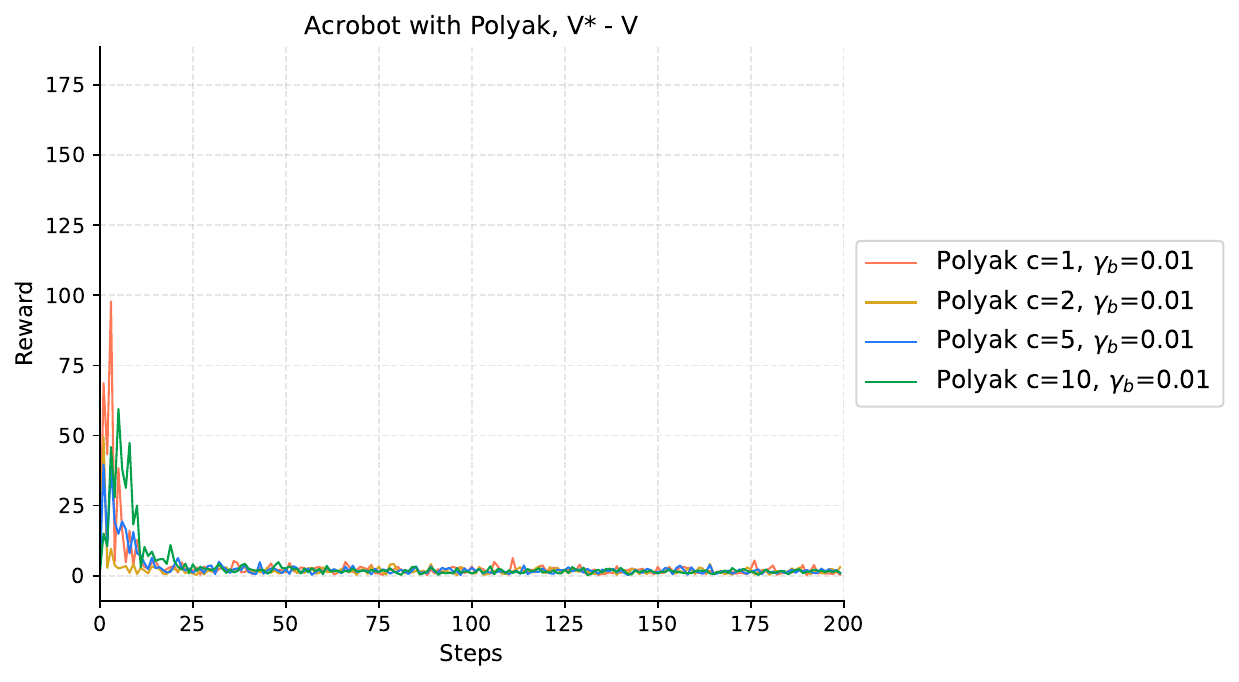}
    \caption{Value of $V^*-V$ in Polyak method in Acrobot.}
\end{figure}

\begin{figure}[H] \centering
    \includegraphics[width=0.45\textwidth]{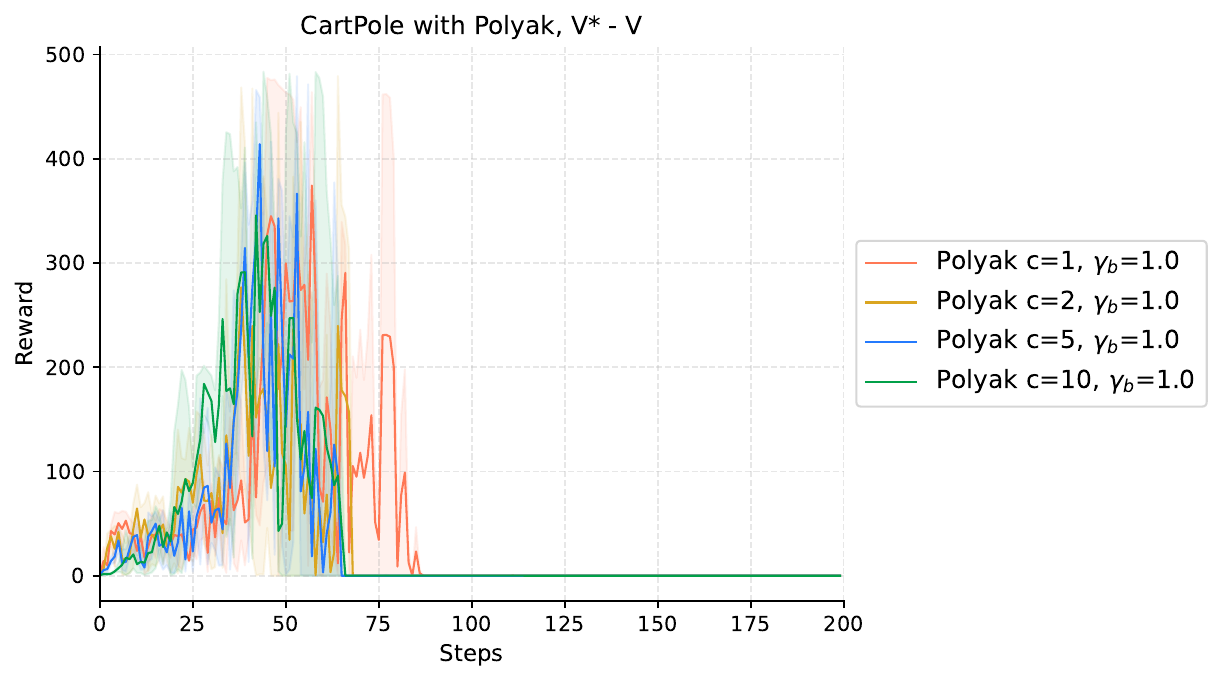}
    \includegraphics[width=0.45\textwidth]{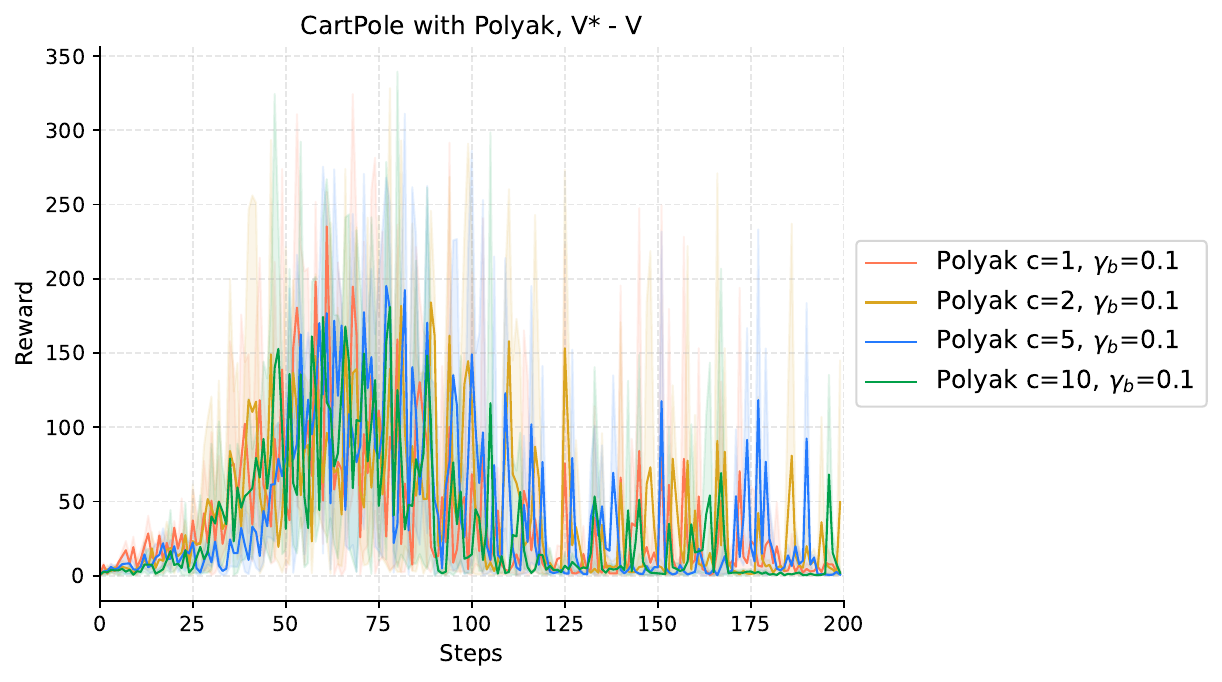}
    \\
    \includegraphics[width=0.45\textwidth]{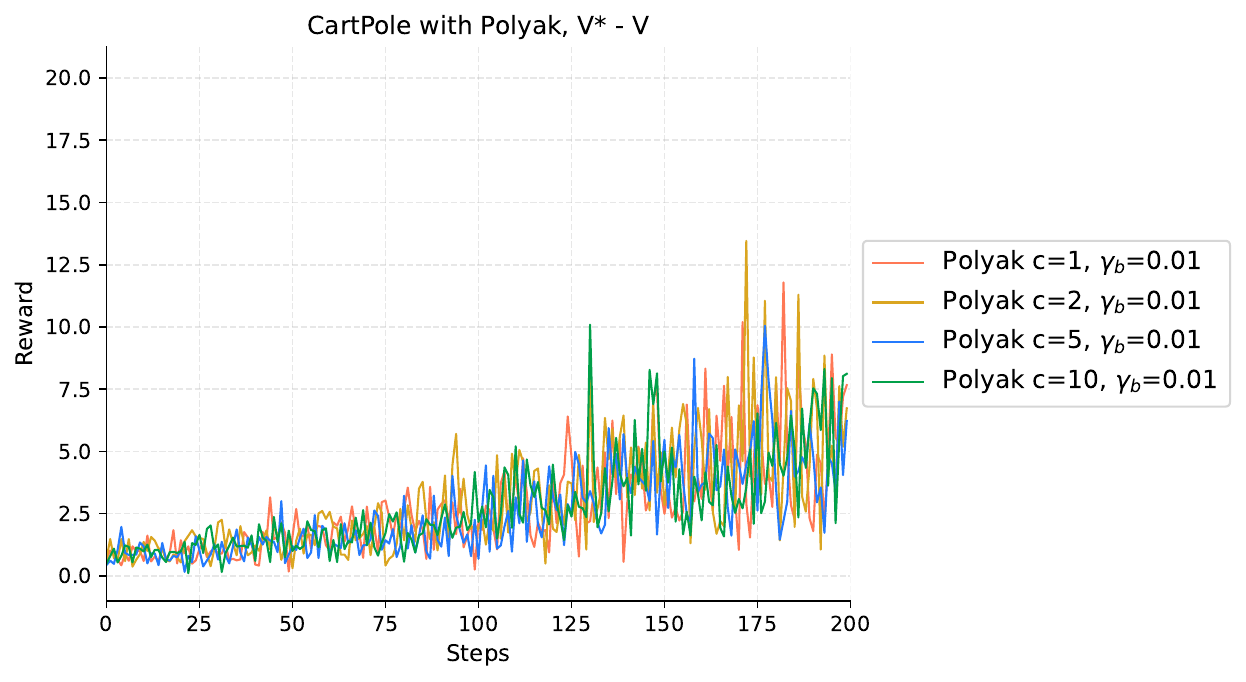}
    \caption{Value of $V^*-V$ in Polyak method in CartPole.}
\end{figure}

\begin{figure}[H] \centering
    \includegraphics[width=0.45\textwidth]{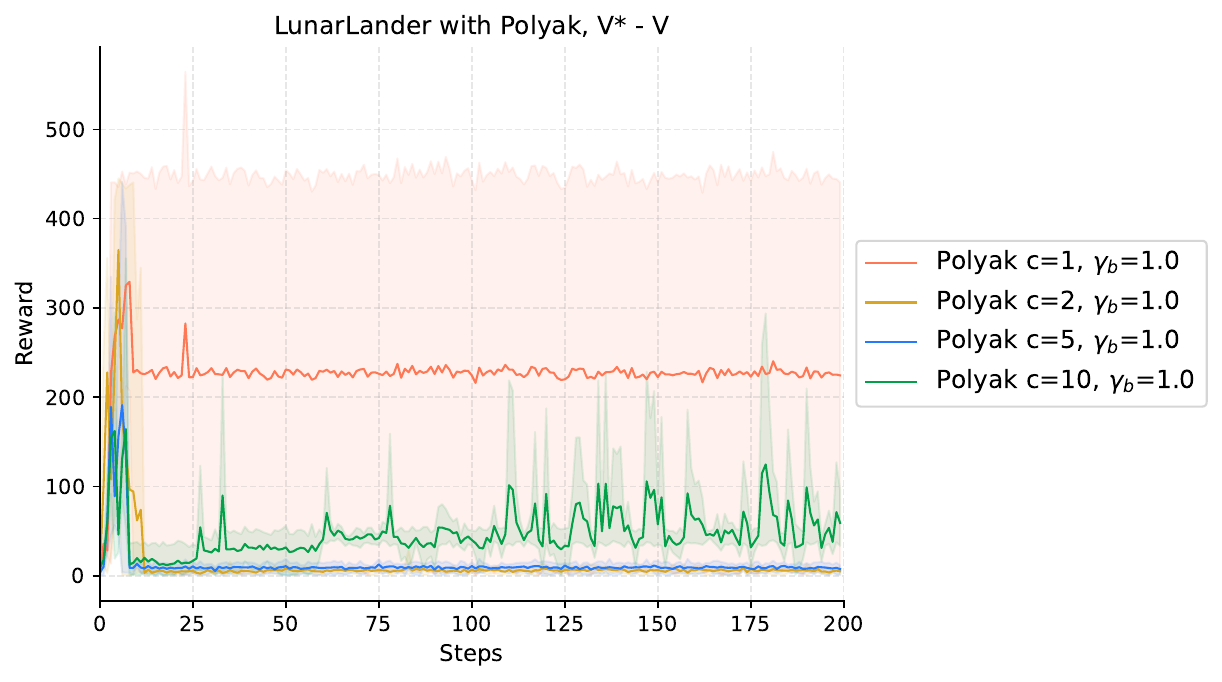}
    \includegraphics[width=0.45\textwidth]{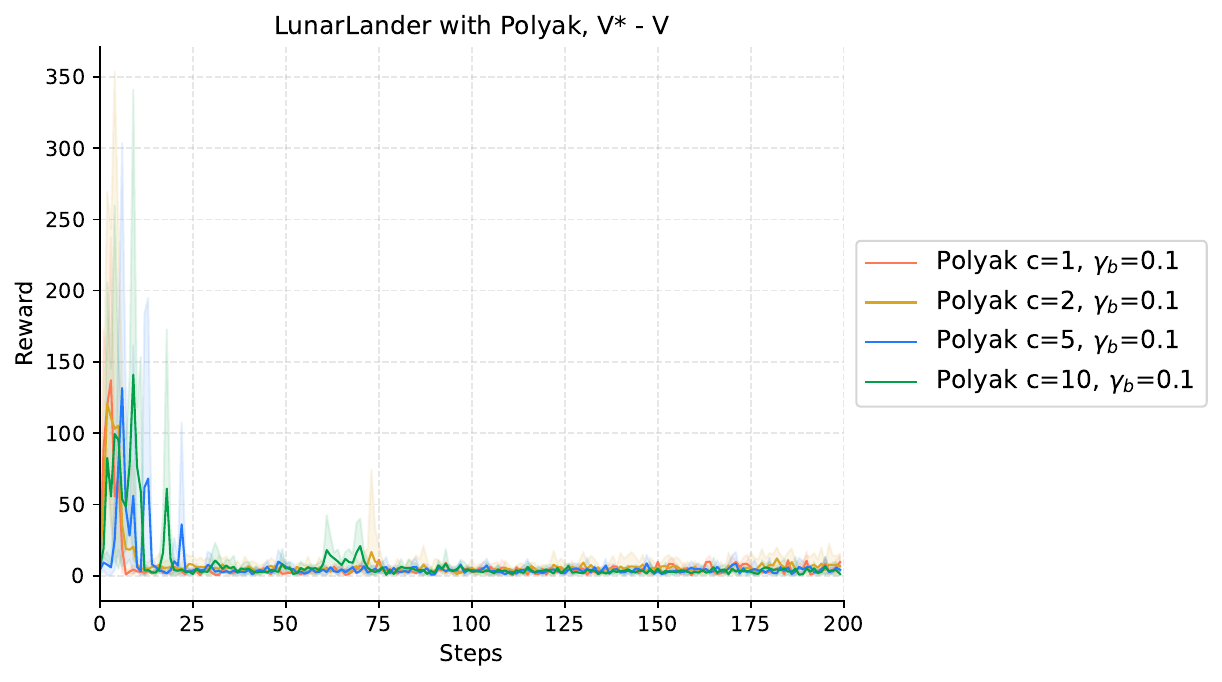}
    \\
    \includegraphics[width=0.45\textwidth]{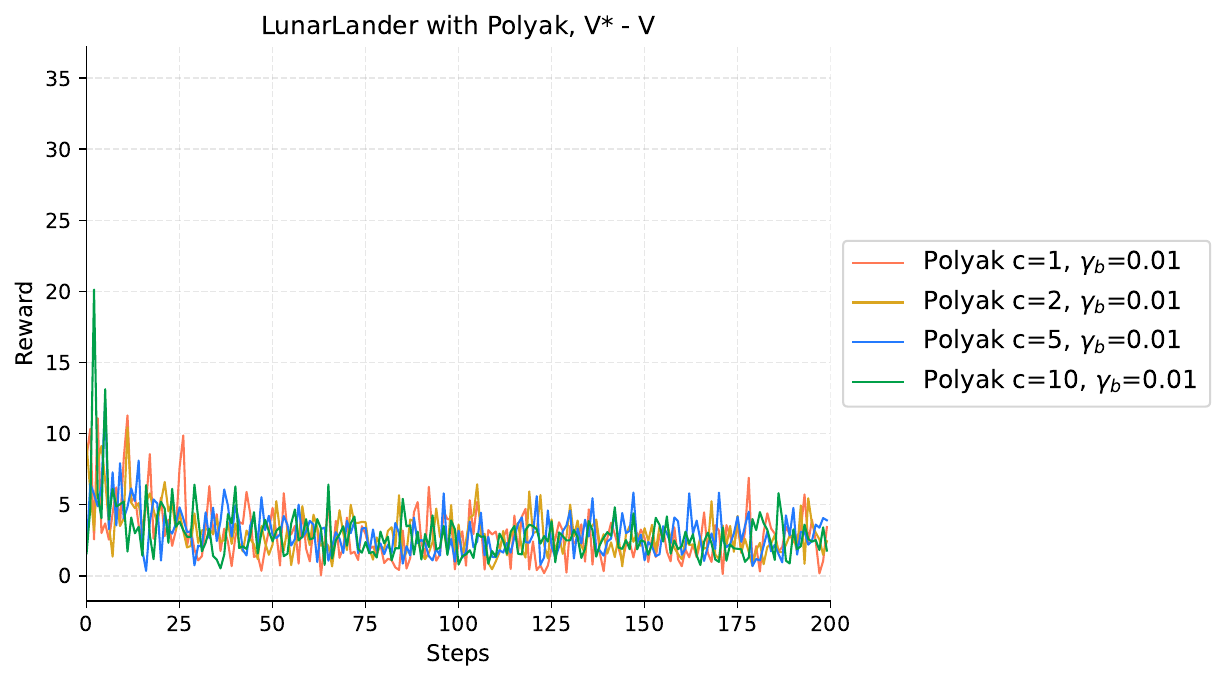}
    \caption{Value of $V^*-V$ in Polyak method in LunarLander.}
\end{figure}

\end{document}